\pdfoutput=1

\documentclass[preprint,12pt]{elsarticle}

\usepackage{times}
\usepackage{epsfig}
\usepackage{graphicx}
\usepackage{amsmath}
\usepackage{amssymb}
\usepackage{url}

\usepackage[pagebackref=true,breaklinks=true,letterpaper=true,colorlinks,bookmarks=false]{hyperref}

\usepackage{graphicx}
\usepackage{color}
\usepackage{booktabs}
\usepackage{multirow}

\usepackage{array}
\newcolumntype{L}[1]{>{\raggedright\let\newline\\\arraybackslash\hspace{0pt}}m{#1}}
\newcolumntype{C}[1]{>{\centering\let\newline\\\arraybackslash\hspace{0pt}}m{#1}}
\newcolumntype{R}[1]{>{\raggedleft\let\newline\\\arraybackslash\hspace{0pt}}m{#1}}

\usepackage{xspace}
\makeatletter
\DeclareRobustCommand\onedot{\futurelet\@let@token\@onedot}
\def\@onedot{\ifx\@let@token.\else.\null\fi\xspace}

\def\eg{e.g\onedot} 
\def\ie{i.e\onedot}

\def\etal{et al\onedot}

\newcommand{\argmax}{\arg\!\max}

\journal{arXiv}

\begin{document}

\begin{frontmatter}



\title{Searching Scenes by Abstracting Things}

\address[label1]{University of Amsterdam, Science Park 904, Amsterdam, the Netherlands}
\cortext[cor1]{Corresponding author}
\address[label2]{Delft University of Technology, Mekelweg 4, Delft, the Netherlands}
\address[label3]{Qualcomm Research Netherlands, Science Park 400, Amsterdam, the Netherlands}

\author[label1]{Svetlana Kordumova\corref{cor1}}
\ead{svetlana.kordumova@gmail.com}
\author[label2]{Jan C. van Gemert}
    \author[label1,label3]{Cees G. M. Snoek}
\author[label1]{Arnold W. M. Smeulders}


\begin{abstract}

In this paper we propose to represent a scene as an abstraction of
``things''. We start from ``things'' as generated by modern object
proposals, and we investigate their immediately observable
properties: position, size, aspect ratio and color, and those
only. Where the recent successes and excitement of the field lie
in object identification, we represent the scene composition
independent of object identities. We make three contributions in
this work. First, we study simple observable properties of
``things'', and call it things syntax. Second, we propose
translating the things syntax in linguistic abstract statements
and study their descriptive effect to retrieve scenes. Thirdly, we
propose querying of scenes with abstract block illustrations and
study their effectiveness to discriminate among different types of
scenes. The benefit of abstract statements and block illustrations
is that we generate them directly from the images, without any
learning beforehand as in the standard attribute learning.
Surprisingly, we show that even though we use the simplest of
features from ``things'' layout and no learning at all, we can
still retrieve scenes reasonably well.

\end{abstract}

\end{frontmatter}


\section{Introduction}

In general, scenes provide the context by which objects receive
their meaning. A picture of the \emph{sea} makes the large pile in
the middle a likely candidate to be an \emph{iceberg} or an
\emph{oil tanker}. And reversely, the understanding of scenes can
be derived from the objects in the scene. The types of objects may
be important, such as a \emph{cow} and another \emph{cow} and a
\emph{tree} to denominate the scene as a \emph{meadow}. With the
current achievements of object recognition
\cite{Everingham15,ILSVRC15} some objects may contribute much to
the recognition of the
scene~\cite{olivaIJCV01spatialEnvelope,objbankIJCV13}. Other
objects may contribute little or noting, like a
\emph{handkerchief} or a \emph{smartphone} as they may occur in
any scene. Where the recent successes and excitement of the field
lie in object identification \cite{Everingham15,ILSVRC15}, in this
paper we argue that next to object types, there is another source
of information contributing to what makes a scene.

\begin{figure}[t!]
\centering
\begin{tabular}{@{}c@{\hspace{5px}}c@{}}
\includegraphics[width=0.44\linewidth, height=4cm]{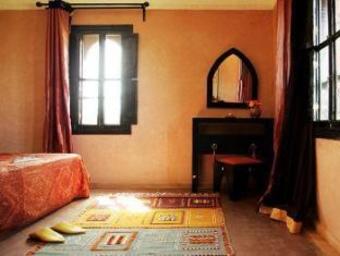} &
\includegraphics[width=0.44\linewidth, height=4cm]{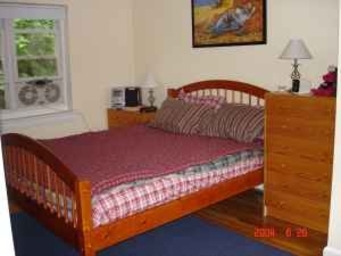}  \\
(a) & (b) \\
\includegraphics[width=0.44\linewidth, height=4cm]{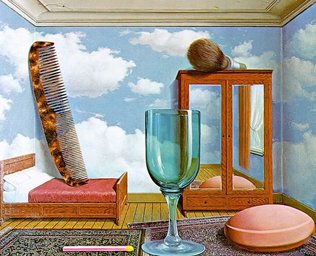} &
\includegraphics[width=0.44\linewidth, height=4cm]{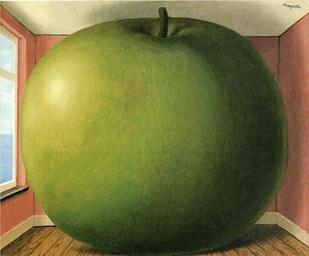} \\
 (c) & (d)
\end{tabular}
\caption{(a-b): The scene type does not change when particular
object are replaced by others in a similar semantic object-class.
(c-d): Scene semantics are violated, not because the objects do
not belong but because they deviate from their typical shape, size
and placement in the scene. \small{Images (c-d) \small{\copyright}
R\'ene Magritte (fair-use).} } \label{fig:BedroomMagritte}
\end{figure}

We note on the basis of cognitive
experiments~\cite{schyns1994blobs} that object composition
provides by itself a clue for the recognition of scenes. The
reference shows that humans can recognize scenes even when
individual objects are reduced to blobs that only retain their
size, aspect ratio and position. As a consequence, when position
and size are violated a scene may appear disorganized, see
Figure~\ref{fig:BedroomMagritte}. By violating these basic
properties, recognizing objects in scenes by humans resulted in
reaction time and accuracy deficit
\cite{biederman1982relationalViolations}. There are reasons why
the scene has a specific spatial layout. Objects obey the laws of
physics. They must be supported by a horizontal surface. Two
objects can not occupy the same physical space. Next to physical
reasons, objects also have a certain semantic likelihood of
appearing in a particular scene \cite{statsHighLevScnCnxGreene}.
Imagine going to a friend's housewarming party, you have never
seen the house before, but you will not be surprised to see a
coffee table next to the sofa in the living room, or framed
pictures on the walls, and you will go look under the sink for
disposing the waste. Next to position, object size is dependent of
cultural habits, human size and purpose
\cite{delaitreECCV12sceneSemanticsPeopleObservation}. The purpose
of a sofa is for sitting so it needs to accommodate a person,
whereas the purpose of a cup is to be hold in hands and
accordingly it is designed smaller. Object size also relates to
scene depth~\cite{torralba2002depth}, which in turn affects the
number of objects ~\cite{Biederman1981}. All of this provides
ample motivation to study the compositional layout of objects in a
scene.

\begin{figure*}
\centering
\includegraphics[width=\linewidth, trim = 0mm 5mm 0mm 25mm, clip=true]{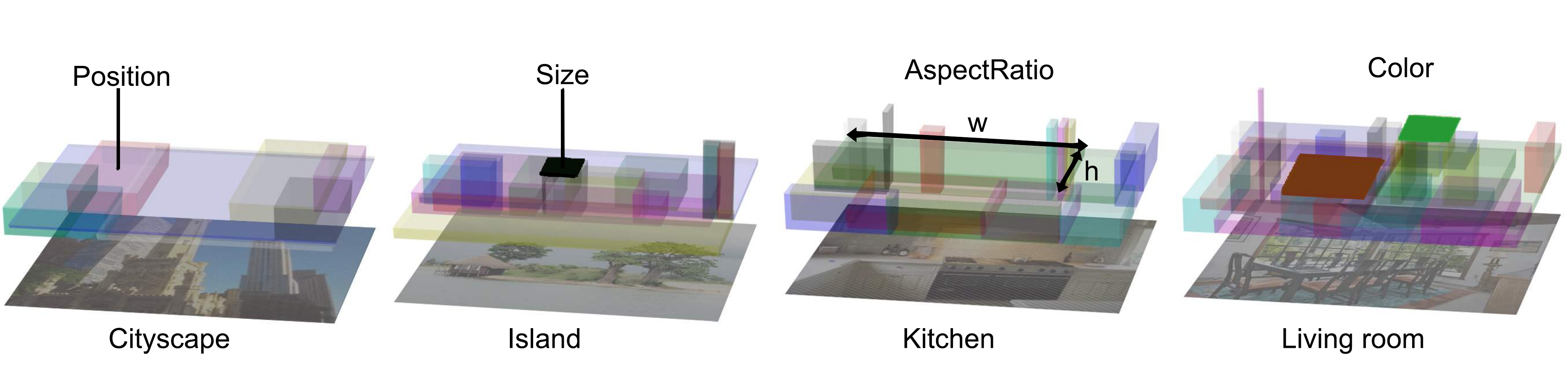}
\caption{Examples of things properties: \emph{position},
\emph{size}, \emph{aspect ratio} and \emph{color}, when things are
visualized as cuboids over the image.} \label{fig:ObjProperties}
\end{figure*}

We have found that the removal of all object identities, and
calling all objects in the scene ``things'', reveals a preference
for a scene-type specific composition of things in a scene. A
\emph{theater} and a \emph{marching band}, a \emph{soccer crowd}
and a \emph{forest} will all demonstrate a certain spatial
regularity of the things they are made of, albeit a spatial
regularity of a different kind each. The interior of a sleeping
room will usually demonstrate a limited number of spatial layouts,
which contribute to the recognition of the scene. In this paper,
we start from ``things'' as generated by modern object proposals,
referred to as objectness~\cite{alexe2012measuring},
PRIM~\cite{manenICCV13objectProposals}, or selective
search~\cite{uijlingsIJCV13selectiveSearch}. We note that some
works have referred to a thing only if it is an ``object that has
a specific size and shape'', and to stuff for ``material defined
by a homogeneous or repetitive pattern of fine-scale properties,
but has no specific or distinctive spatial extent or shape''
\cite{Forsyth:CSD-96-905, stuff-things-HeitzK08}. In our work we
name all existing objects, parts and stuff under ``things''. We
investigate image representations composing the ensemble of these
things, on the basis of their immediately observable features:
position, size, aspect ratio and color, and those only, see
Figure~\ref{fig:ObjProperties}. We could refer to the
representation as sceneness, SRIM, or selective layout, but we
prefer to use \emph{things syntax}.

To relate visual information with linguistic meaning is another
challenging and open area of research. Linguistic descriptions are
usually generated using object identities \cite{objbankIJCV13,
farhadi2009describing, babytalkKulkarni11} or attributes
\cite{Patterson2012SunAttributes}. In this paper we take a
different approach and generate descriptions with abstract
statements only from the immediately observable features of
``things'', for example ``Green small squared thing at top
middle'' or ``Blue large wide thing at top right''. Whereas
\cite{objbankIJCV13, farhadi2009describing, babytalkKulkarni11,
Patterson2012SunAttributes, abstractPAMI} use manual annotations
and require learning to generate linguistic descriptions, we
generate linguistic statements from the image itself. We
investigate the effectiveness of the abstract statements to
retrieve scenes without the need for examples.

Furthermore, we investigate querying the things syntax with
abstract block illustrations, as presented in
Figure~\ref{fig:AutomaticBlocksStats}, inspired by Mondrian's
neoplastic style~\cite{mondriaanDeStijl}. Mondrian studies the
order of the most abstract forms to designate a feeling for the
scene as a whole. The painting "Victory Boogie Woogie" is a good
example of expressing the crowdedness of a joyful city. In our
work, different from \cite{abstractPAMI}, we remove object
identities and use only abstract block illustrations that preserve
the immediate observable properties of things, see
Figure~\ref{fig:ObjProperties}. We study the effectiveness of
block illustrations to search scenes.

We make three contributions in this work. First, on the basis of
the above motivation, we approximately localize the things in a
scene as the output of objectness \cite{alexe2012measuring},
selective search \cite{uijlingsIJCV13selectiveSearch}, and PRIM
\cite{manenICCV13objectProposals}, and study the properties of
their immediate observables without proceeding to identify these
things by type. Second, we study abstraction of the things syntax
by translating it into abstract statements and their effectiveness
to query scenes. And thirdly, we study abstraction of scenes into
block illustrations of things to search among different types of
scenes.

\section{Related Work}

\textbf{Scene representation.} To represent and to recognize
scenes more approaches have been proposed than we can cover here.
We make no attempt to be complete. Instead, we summarize the main
research strands which developed from low level statistics,
\eg,~\cite{olivaIJCV01spatialEnvelope,
wuPAMI11centrist,sanchezIJCV13fisherVector,deepSceneLazebnikECCV14},
through mid-level unnamed discriminative regions, \eg,
\cite{guCVPR09recognition,gouldICCV09decomposingScene,doersch2013mid,
junejaCVPR13blocksThatShout} to high-level representations
consisting of objects identity scores, \eg,~\cite{objbankIJCV13,
classemesPAMI} or attributes \cite{Patterson2012SunAttributes,
lampert2013attribute}. Each of these are successful in their own
right and for their intended purpose. In this paper we follow a
different path to search the categories of scenes and while doing
so we demonstrate that for scene recognition there is another
informative cue besides appearance features and object types.

\textbf{Scene statistics}. We take as inspiration the work of
Greene \cite{statsHighLevScnCnxGreene}, who investigates object
statistics like object density, unique object density, mean object
size, and variability of objects, to discriminate scenes. These
statistics are calculated from manually annotated objects in scene
images, with known object types. Interestingly, Greene shows that
this ensemble of statistics is sufficient for above-chance scene
recognition. Motivated by these results, we investigate
immediately observable properties of nameless things. Thanks to
the modern object proposals methods \cite{alexe2012measuring,
uijlingsIJCV13selectiveSearch, manenICCV13objectProposals}, we can
benefit from automatically suggested locations of things. We
investigate how close do the immediate observable properties of
automatically generated object proposals come to the properties of
manually annotated objects, and compare their discriminative power
for retrieving scenes.

\textbf{Object proposals}. We use the recent achievements of the
localization of objects to find things in a scene
~\cite{alexe2010object,endresECCV10categoryIndependetObjProp,
manenICCV13objectProposals,rahtu2011learning,
uijlingsIJCV13selectiveSearch, GeodesicObjProp,
ZitnickDollarECCV14edgeBoxes}. The goal of these methods is to
find locations in an image that have a high likelihood to contain
an object. By being efficient in suggesting the most likely
locations, these methods allow more computation time to be spent
on feature representations and classifying their identify for
object detection, which has led to great successes. Very often the
proposed locations are around an object part or around some
texture shapes. Since we are considering things, there is no
longer an objective to identify an object. An object part can
often be considered as an object in its own right, for example
buildings in a cityscape have windows, apartments and individual
bricks. In natural scenes, this recursive fragmentation follows
certain rules biasing the statistics of size towards a Weibull
distribution~\cite{geusebroek2005six}. Starting from
~\cite{alexe2010object,uijlingsIJCV13selectiveSearch,manenICCV13objectProposals}
we consider their output locations in a form of a bounding box
over an image to hold a thing, and we refer to them as things
proposals in the rest of the paper. We gladly use their valuable
output, right or wrong, without proceeding to classify the content
of their boxes into an object type, as these methods are designed
for. In this paper we are satisfied with the position, size,
aspect ratio and color of the proposed boxes to represent the
ensemble of just the things.

\textbf{Semantic scene representation}. Predicting written
descriptions from visual information is an interesting and
challenging problem \cite{FarhadiPicTellStorry,
babytalkKulkarni11, farhadi2009describing, RelAttr-ParikhG11,
im2text-Ordonez2011, BergAttributesWeb, CorpusSentenceYang2011}.
All these works use a variety of approaches, like generating
semantic sentences relying on object detectors
\cite{objbankIJCV13, babytalkKulkarni11}, relying on semantic
attributes \cite{Patterson2012SunAttributes,
farhadi2009describing, BergAttributesWeb, RelAttr-ParikhG11},
using a large corpus to extract written descriptions
\cite{FarhadiPicTellStorry, im2text-Ordonez2011}, language
statistics \cite{CorpusSentenceYang2011} or generating verbs by
looking into spatial relationships of objects
\cite{babytalkKulkarni11}. Notable efforts have also been done in
creating datasets with semantic descriptions and sentences
\cite{LinMBHPRDZ14, Rashtchian2010}. In this paper we take a
different approach by removing objects, attributes and verbs from
written descriptions. We describe a scene by mapping properties of
things to adjectives and adverbs, for example, \emph{Green small
squared thing at top middle} or \emph{Blue large wide thing at top
right}.

\textbf{Zero-shot recognition}. The benefit of having high-level
written descriptions of images is that they enable humans and
computers to communicate with natural language. One popular
application is zero-shot recognition as pioneered by Fahradi
\emph{et al.} \cite{farhadi2009describing} and Lampert \emph{et
al.} \cite{lampert2013attribute}. In a zero-shot setting, there
are no available category labels on images to learn from, instead
human descriptions of zero-shot classes are gathered and matched
against detected scores of attributes or object detectors in test
images. This setting requires gathering annotations and learning
models of attributes \cite{RelAttr-ParikhG11,
lampert2013attribute, farhadi2009describing} or objects detectors
\cite{JainICCV15, objbankIJCV13}, from the same scope as the test
classes. For example animal attributes can not be used to
recognize vehicles. Images from the zero-shot classes are usually
shown to humans in the description process, since if a person has
never a seen an \emph{archeological excavation} scene, it would be
difficult to provide attributes for it. In this paper we remove
all scope-dependent object identities and attributes, and
represent scenes as an ensemble of ``things''. We pose the
question, \emph{Can we retrieve a scene with only abstract
statements of immediately observable things properties?}. An
important benefit of the abstract statements is that they are
generated directly from images, without the need to learn any
attribute or object detectors beforehand.

\textbf{Visual abstraction}. Zitnik \emph{et al.}
\cite{abstractPAMI} have investigated the potential of using
abstract images to study high-level semantics, like semantically
important features, relations between saliency-memorability of
objects, and mapping of sentences to abstract scenes. A dataset of
abstract images is collected by asking users to draw images with
clip art objects of children, trees, animals, food, toys etc. The
authors number four benefits of abstract images: 1) they remove
the reliance on noisy low level object and attribute detectors 2)
they avoid tedious hand labeling on images, 3) they allow for
direct study of high level semantics, and 4) allow to
automatically generate sets of semantically similar images. We
stand inspired and propose to investigate abstract images from a
different perspective, using nameless things. Different from
\cite{abstractPAMI} we remove all object identities and create
abstract block illustrations capturing the position, ratio, size
and color of things in a scene. Abstract images have also been
investigated as a modality for zero-shot recognition by Antol
\emph{et al.} \cite{Antol2014}. The authors argue that visual
modality is needed because some images are not easy to be
described in semantic terms, like images of interactions between
people, which is also the topic of their work. Since we are
interested in scene images, we note that describing scenes in
semantic terms is also difficult in a number of situations. For
example, when children between the age of 5 to 6 are asked to
describe a memory, they show more accurate information by drawing
the scene than when they describe it semantically
\cite{childrenDrawingMemory}. Also there are scenes where object
identities are unknown, but their ensemble of things is well
defined, like cosmic and microscopic scenes, architectural
inspiration or abstract paintings. We investigate searching scenes
by block abstractions.

In this paper, rather than using sophisticated scene
representations, we evaluate simple features from the composition
of ``things'' in the scene, from which we define query
representations based on simple linguistic statements and simple
block illustrations.



\section{Things syntax}

Starting from three object proposal methods
~\cite{alexe2010object, uijlingsIJCV13selectiveSearch,
manenICCV13objectProposals}, we consider their output locations in
an image to hold a thing. The output locations are bounded with a
box and we refer to them as windows. We use the bounding box
window $w$ to calculate the thing properties. For
\textbf{horizontal position} and \textbf{vertical position} we use
the center coordinates $w_x$ and $w_y$ of its window $w$. Thing
\textbf{size} is approximated with the window width and height
$w_s = w_ww_h$ whereas shape is measured by the \textbf{aspect
ratio}
\begin{equation}
w_r = \left\{
\begin{array}{ll}
0.5 (w_w/w_h) & \quad \mbox{if $o_w \ge o_h $ }; \\
0.5 (w_h/w_w + 1) & \quad \mbox{if $o_w < o_h $} ,
\end{array} \right. \label{eq:ratio}
\end{equation}
which has a value of $0.5$ for square objects, a value between $0
\le w_r< 0.5$ for tall objects and a value $0.5<w_r<1$ for broad
objects. We also measure the dominant \textbf{color} $w_c$ of the
thing window with eleven basic colors as defined by
\cite{colornaming2009}.

The accidental image sensor resolution influences thing window
size and position. If an image is scaled by a factor $f$, the
windows will be scaled accordingly. The window center will move to
$(fw_x, fw_h)$ and the window size will become $fw_wfw_y$. Window
aspect ratio is invariant to the image resolution since
$\frac{fw_w}{fw_h} = \frac{w_w}{w_h}$. We obtain resolution
invariance for window position and size by normalizing with the
image width $I_w$ and height $I_h$ as $\frac{w_x}{I_w}$
since $\frac{fw_x}{fI_w} = \frac{w_x}{I_w}$. We also investigated
translation and scale invariance of thing windows, but we found
that adding these invariants does not affect the discriminative
potential of things windows.

Each thing window is represented by: horizontal position, vertical
position, size, ratio and color, resulting in a $1x5$ dimensional
vector $\mathbf{w} = [w_x,w_y,w_s,w_r,w_c]$. The ensemble of all
windows properties in an image forms the things syntax. If an
image has $n$ thing windows, represented as vectors
$\{\mathbf{w}_i\}_{i=1}^{n}$, then the image things syntax is
represented by stacking all windows vectors in a matrix

\begin{align}
    W^{nx5} &= \begin{bmatrix}
                \mathbf{w}_{1} \\
                \mathbf{w}_{2} \\
                \vdots \\
                \mathbf{w}_{n}
                \end{bmatrix} =
    \begin{bmatrix}
                w_{1x},w_{1y},w_{1s},w_{1r},w_{1c}\\
                w_{2x},w_{2y},w_{2s},w_{2r},w_{2c}\\
                \vdots \\
                w_{nx},w_{ny},w_{ns},w_{nr},w_{nc}
                \end{bmatrix}.
  \end{align}

\subsection{Query by abstract linguistic statements}\label{sec:statements}

\begin{figure*}[t!]
\centering
\includegraphics[width=0.99\linewidth, height=4.2cm]{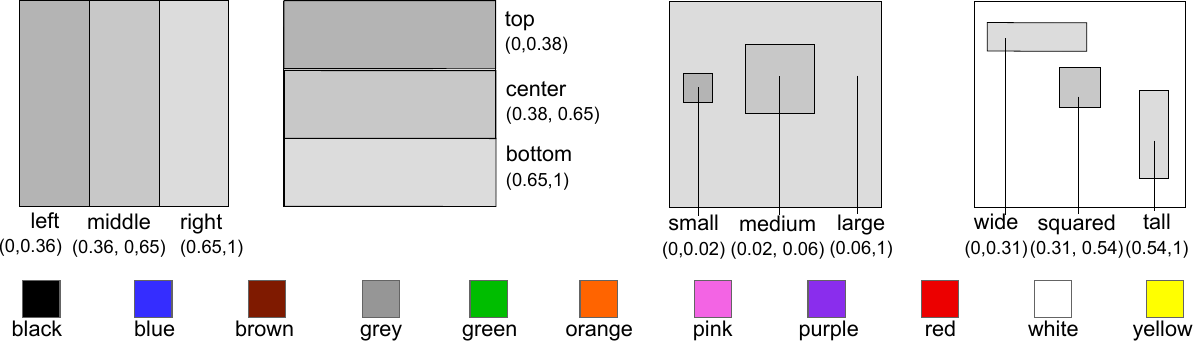}
\caption{Binning boundaries for quantization of the continuous
things properties: horizontal position, vertical position, size
and shape, to a single word like \emph{left}, \emph{top},
\emph{small} and \emph{wide}. We quantize each thing property
separately and combine them in one histogram representation later.
We use eleven colors to represent the most dominant color in a
thing window. Thus for each thing an abstract statement like
\emph{Green small wide thing at top left} can be created.}
\label{fig:RecountingBoudries}
\end{figure*}

To formalize a scene image in linguistic statements, the things
syntax needs grounding in natural language. Such a grounding can
be obtained by quantizing the window properties of things to human
understandable statements. Since we purposely ignore object names,
we disregard nouns, and map window properties to adjectives
(small, green, tall) and adverbs (left, middle, right). We chose
three levels since humans tend to use three denotations for object
properties. For example, horizontal position can be quantized to
the three words (left, middle, right) and size can be mapped to
(small, medium, large). In Figure~\ref{fig:RecountingBoudries} we
summarize the set of rules we define to translate the horizontal
position, vertical position, size and shape into a things
vocabulary composed of adverbs and adjectives. We also map the
color of the things windows on one of the eleven dominant color
names \cite{colornaming2009}.

\textbf{Histogram representation}. Mapping a continuous property
of a thing window, such as its x-position $w_x$, to a single word
such as ``left'' requires setting binning boundaries for
quantization. We quantize each thing property separately and
combine them in one histogram representation later. Each bin in
the histogram represents an occurrence value for one combination
of things properties, and one sentence like ``Green large tall
thing at bottom left'' can be created. A straightforward solution
for binning is to split the axis of each window property value in
equal bins $[0,\frac{1}{3},\frac{2}{3}]$. However, things windows
may not be uniformly distributed, and thus frequent words may bias
the description. On a separate holdout set we calculate simple
statistics of things properties to obtain an equal probability of
word occurrence. The holdout set can be any scene images,
downloaded from internet for example, or by ignoring the
annotations on an existing scene dataset, since we calculate
statistics directly from the images and we do not need
annotations. With a histogram $h$ of three bins over the training
images, the word bin boundaries are obtained by splitting the
histogram in three parts with equal probability weight, i.e.,
$\sum_0^{i_1}h(i) = \sum_{i_1+1}^{i_2}h(i) = \sum_{i_2}^{1}h(i) $,
where $i_1$ and $i_2$ indicate the bin separator. In
Figure~\ref{fig:RecountingBoudries} we show word boundaries
obtained in this manner on a holdout set. With binning boundaries
we can quantize properties to words, and thus convert a set of
things windows to a probability distribution over statements. If
for horizontal position, vertical position, size and ratio we use
3 words, and for color we use 11 words \cite{colornaming2009}, the
histogram representation of the probability distribution over all
statements combinations will be 3x3x3x3x11 dimensional.

\textbf{Query by abstract statements}. Since the abstract
statements are in a human understandable form, they allow us to
search unseen scene classes. For example, if the unseen class is a
bedroom, the formal statements allow, in principle, for a
description of this class with ``wide big brown thing in the
center, small squared things left and right''. When abstract
statements like this are provided for a scene, we can easily
create a histogram representation. Creating a histogram
representation from abstract statements is a reverse process of
translating the things properties into abstract statements
described above. Each statement stands for the properties of one
thing, and is counted in one bin of the histogram representation.
The value of each bin is a count on how many times the
corresponding statement was used to describe a particular scene.
At test time, we compare the histogram representation of scenes
with the histogram representation of test images to compute
probability score of a test image belonging to a scene class. This
procedure is equivalent to the attribute representation of Lampert
\etal~\cite{lampert2013attribute}. The difference is that for
unknown scenes as representations they use a distribution of an
attributes occurrence, whereas we use a histogram of abstract
statements. This allows us to use the same Direct Attribute
Prediction (DAP) model for ranking test images by probability
scores. The DAP-model assigns a test image to a most likely scene
$z_1,\dots,z_L$ with $\argmax_l p(z=l|x) = \argmax_l
\prod_{m=1}^{M} \frac{p(a^{z_l}_m | x) }{ p(a^{z_l}_m)}$, where
$p(a^{z_l}_m | x)$ is the probability value of the attribute
detector of $a^{z_l}$ applied on the test image $x$, and
$p(a^{z_l}_m)$ is the value of the $m$-th attribute occurring in
scene $z_l$. In our case we have statement representations per
scene instead of attributes. One important difference in this
formulation is that $p(a^{z_l}_m | x)$ in
\cite{lampert2013attribute} is calculated from attribute
detectors, pre-learned on another independent manually annotated
set. We calculate the probability distribution over statements of
a test image $x$ directly from its things properties. In this way
the search process is simplified without the need for any
learning.

\begin{figure}
\centering
\includegraphics[width=\linewidth]{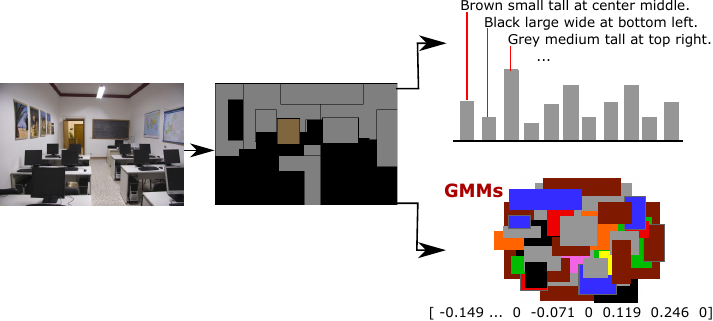}
\caption{Example of creating two things syntax representations.
From a test image, things windows are generated automatically with
a things proposals method. The dominant from eleven pre-defined
colors is preserved from each window. The things properties are
then either quantized into a histogram by binning, or a Fisher
vector is created using pre-calculated GMMs. The histogram
representation is used when searching scenes by abstract
statements, and the Fisher vector representation is used when
searching scenes by block illustrations. }
\label{fig:AnoObjIllustrate}
\end{figure}

\subsection{Query by abstract block illustrations}\label{sec:blockillustrat}

To search by block illustration of a scene, it is not required for
one to remember and draw the exact shape of things. As long as one
can mimic the basic size, form, position and dominant color of
things in the scene.

\textbf{Fisher vector representation}. To represent the block
illustrations and the things syntax of images, we employ the
popular Fisher vector encoding ~\cite{sanchezIJCV13fisherVector}.
On a holdout set $H$, for each scene image $\{I | I \in H\}$ we
compute its things syntax $W_I^{nx5}$. We merge all $\{W_I^{nx5},
I \in H\}$, in one matrix $W_H$ holding all things properties of
the holdout set, and compute Gaussian Mixture Model (GMM)
prototypes of $W_H$. A GMM with $K$ components models the
probability of all things windows properties $w \in W_H$, given
the model $\lambda$ by $P(w|\lambda) = \sum_{i=1}^k t_i g(w|\mu_i,
\Sigma_i)$, where $g$ is the Gaussian function and
$\sum_{i=1}^{k}t_i=1$. For a new block illustration $I$ or image
$I$, it's things properties are encoded with the derivative of the
mean and variance to the GMM
prototypes~\cite{sanchezIJCV13fisherVector} as $\nabla \mu_k \log
g_k( w ) = \gamma_k(w) \left( \frac{w - \mu_k}{\sigma_k^2}
\right)$, and $\nabla \sigma_k \log g_k( w ) = \gamma_k(w) \left[
\frac{ (w - \mu_k)^2}{\sigma_k^3} - \frac{1}{\sigma_k} \right]$,
where $w$ is one row of the things syntax matrix $W_I$, $\mu_k$
denotes the mean, and $\sigma^2_k$ the diagonal of the covariance
matrix $\Sigma_k$ of Gaussian $g_k$, and $\gamma_k(w)$ are the
soft assignment responsibilities of window $w$ to Gaussian $k$.
The final Fisher vector representation of the things syntax is
created as a concatenation of $\nabla \mu_k \log g_k( w )$ and
$\nabla \sigma_k \log g_k( w )$ for each Gaussian prototype $k$.
Simply said, the procedure of the Fisher vector encoding of the
things syntax is equivalent to the Fisher vector encoding of SIFT
vectors \cite{sanchezIJCV13fisherVector}. The only difference in
our case is that instead of using a 128-dimensional SIFT vector
sampled from dense or salient points of an image, we use a
5-dimensional vector of individual thing window properties in an
image. The rest of the procedure is identical.

\textbf{Query by block illustrations}. When block illustrations
$I$ for a scene are available, we can create a Fisher vector
representation of the scene. This representation allows us to
search images of an unseen scene class. We merge all window
properties of block illustrations for a scene $S$ into one things
syntax matrix $W_S$ of the scene. This matrix holds the layout of
things within the scene through its things properties. Following
the procedure described above, we create a Fisher vector
representation from $W_S$. To retrieve test images of an unseen
scene class, we first compute a Fisher vector representation of
the image, and compare it to the scene Fisher vector
representation. Different similarity measures and metric learning
can be adopted to measure the similarity between the Fisher vector
representation of a scene and the image. For now we simply use an
Euclidian Distance to rank the test images. We summarize the
process of creating things syntax representations in
Figure~\ref{fig:AnoObjIllustrate}.

\section{Datasets}\label{sec:datasets}

\begin{figure*}[th!]
\centering
 \scalebox{0.75}{
\begin{tabular}{c@{}c@{}c@{}c@{}c@{}}
& & \textbf{\large Image} & \textbf{\large a) Things proposals} & \textbf{\large b) Annotated things}\\
\cmidrule(lr){3-5}
 \rotatebox{90}{\textbf{\large Football field}} & . &
\includegraphics[width=0.33\linewidth,height=2.8cm]{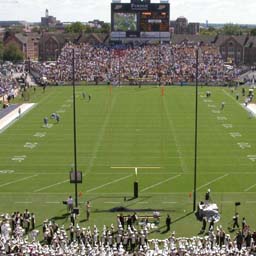}
&
\includegraphics[width=0.33\linewidth,height=3cm,trim = 3.3cm 8.1cm 1.8cm 8.3cm,
clip=true]{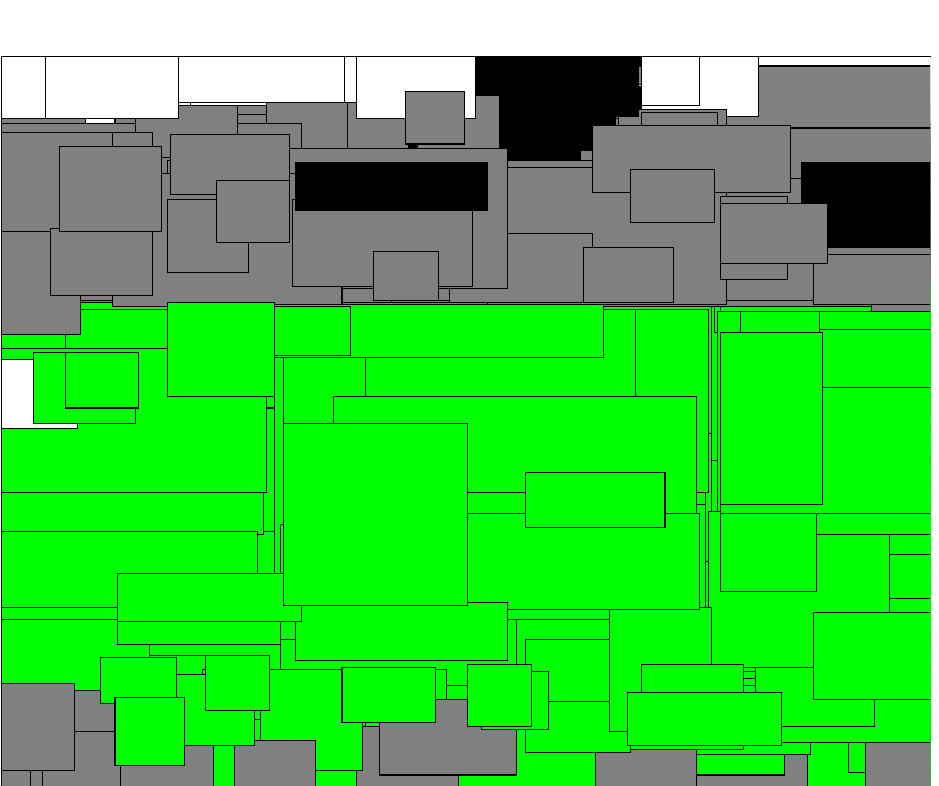}
&
\includegraphics[width=0.33\linewidth,height=3cm,trim = 3.3cm 8.1cm 1.8cm 8.3cm,
clip=true]{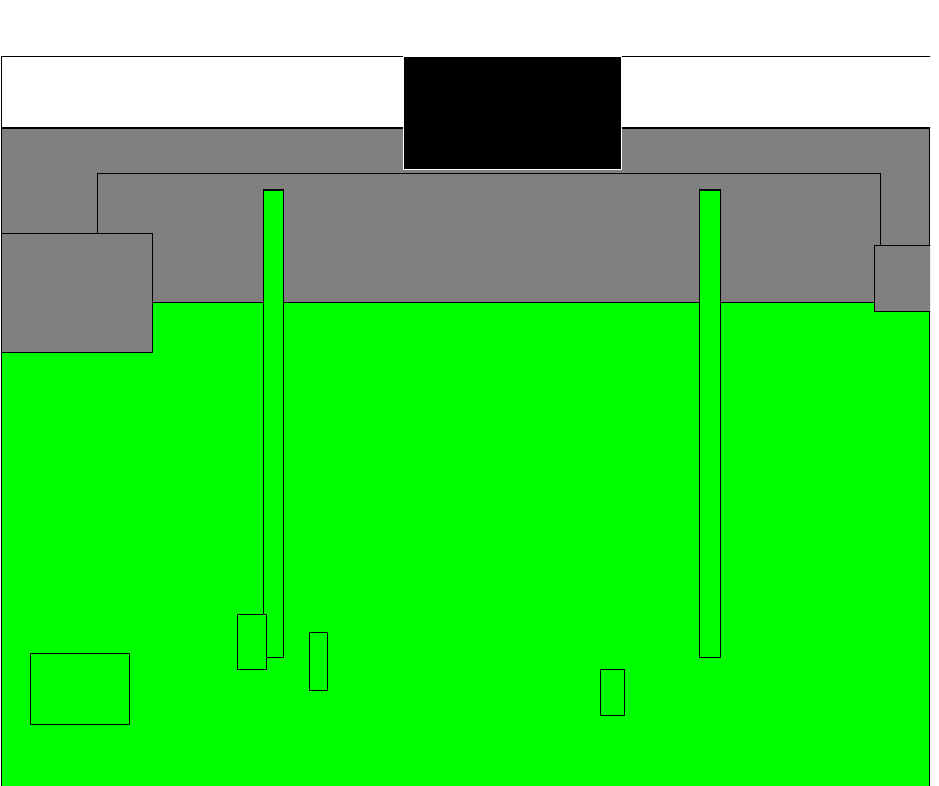}\\
& & & Black medium squared at top middle. & Green large squared at bottom middle. \\
& & & Green large squared at bottom middle. & Green small tall at bottom left. \\
& & & Green medium tall at center right. & Green small tall at bottom right. \\
& & & Green medium tall at center left. & Black medium squared at top middle.\\
& & & Grey medium squared at top left. & Grey large wide at top middle. \\

\cmidrule(lr){3-5}

\rotatebox{90}{\textbf{\large Beach}} & &
\includegraphics[width=0.33\linewidth,height=2.8cm]{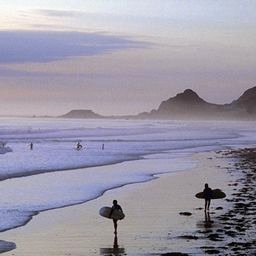}
&
\includegraphics[width=0.33\linewidth,height=3cm,trim = 3.3cm 8.1cm 1.8cm 8.3cm,
clip=true]{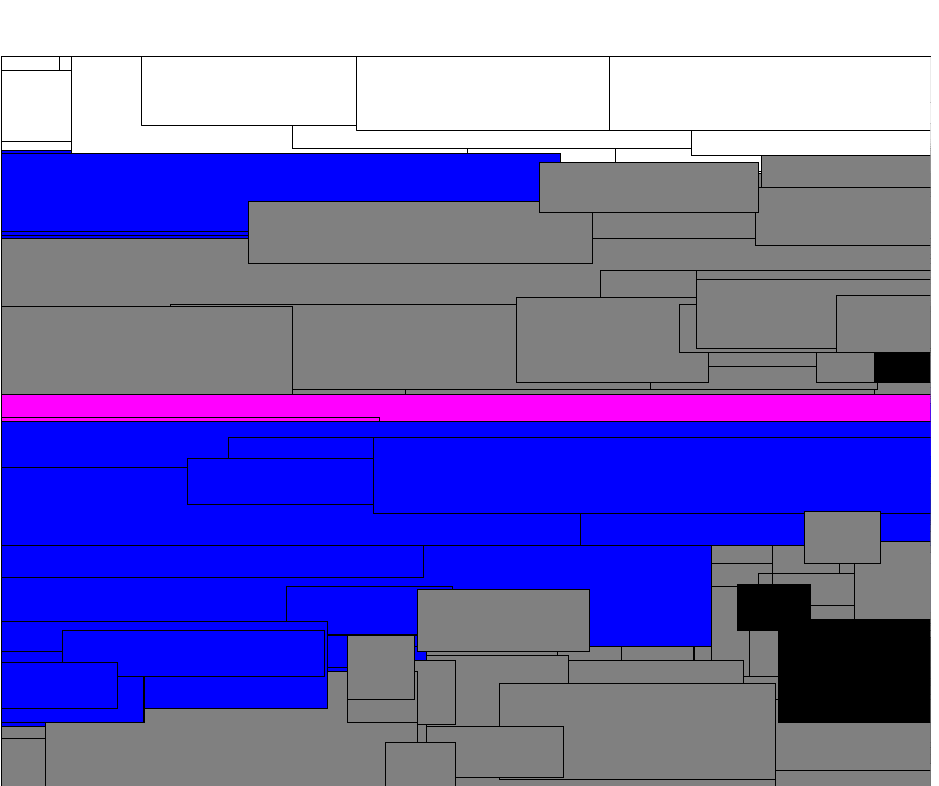} &
\includegraphics[width=0.33\linewidth,height=3cm,trim = 3.3cm 8.1cm 1.8cm 8.3cm,
clip=true]{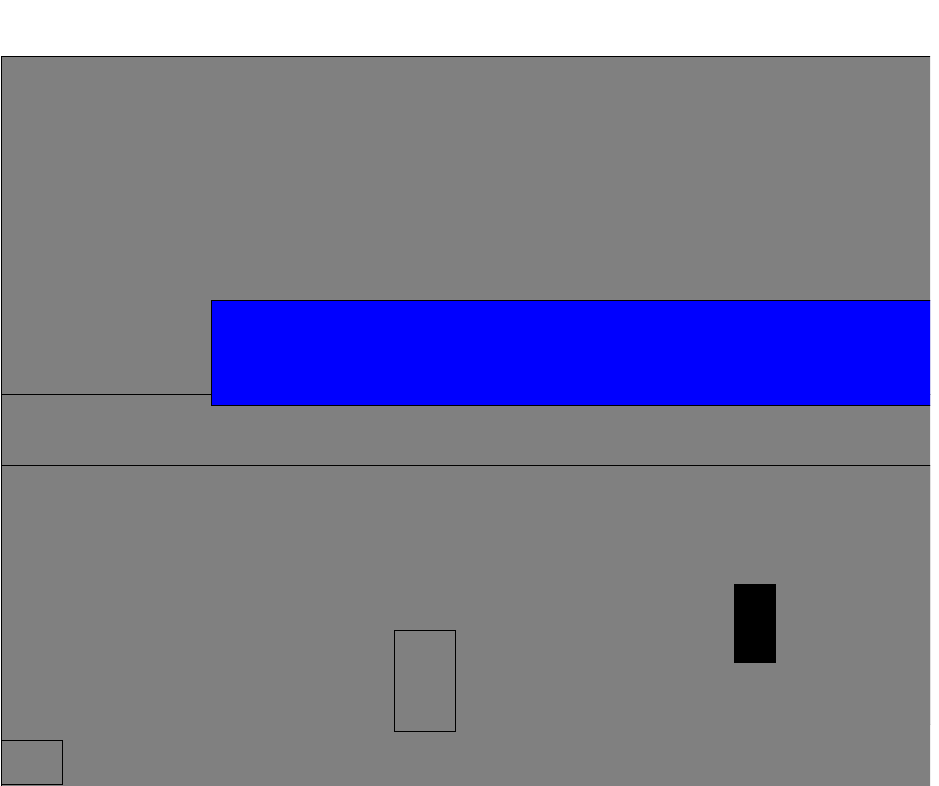}\\
& & & Blue large wide at center. & Grey large wide at top right. \\
& & & Blue large wide at top right. & Grey small squared at bottom left. \\
& & & White small wide at center middle. & Blue large wide at center middle. \\
& & & Grey large wide at top middle. & Grey large wide at bottom middle.\\
& & & Grey large wide at bottom middle. & Black small tall at bottom right. \\

\cmidrule(lr){3-5}

\rotatebox{90}{\textbf{\large Skyscrapers}} & &
\includegraphics[width=0.33\linewidth,height=2.8cm]{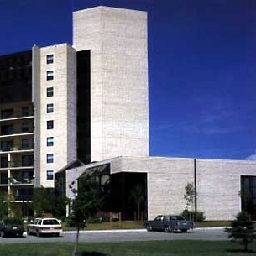}
&
\includegraphics[width=0.33\linewidth,height=3cm,trim = 3.3cm 8.1cm 1.8cm 8.3cm,
clip=true]{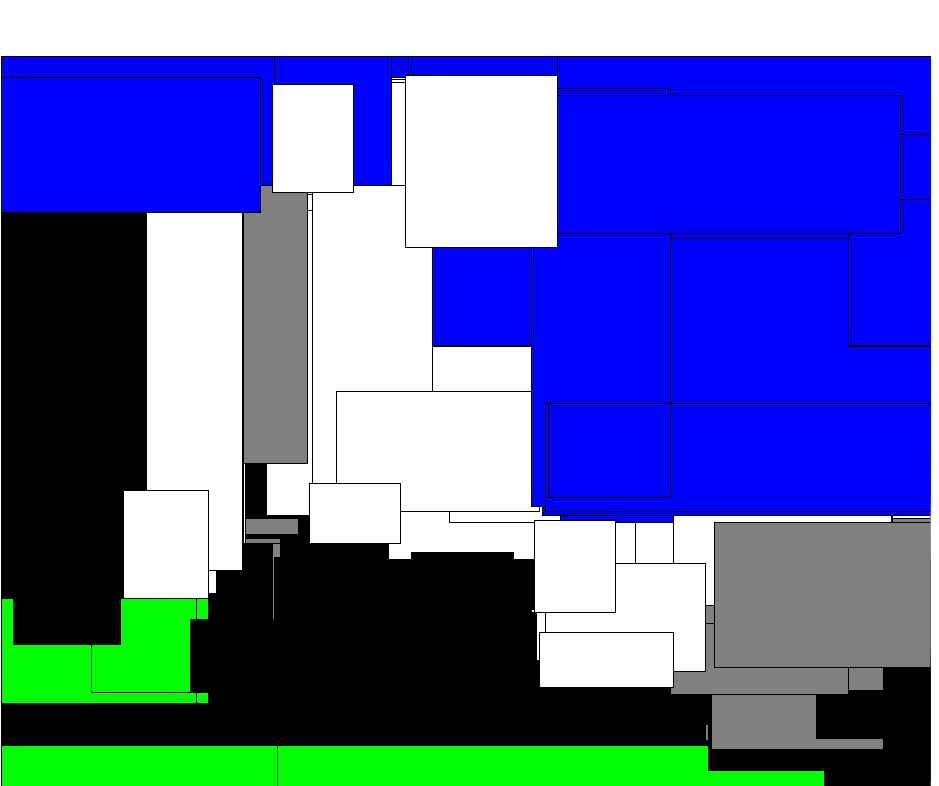}&
\includegraphics[width=0.33\linewidth,height=3cm,trim = 3.3cm 8.1cm 1.8cm 8.3cm,
clip=true]{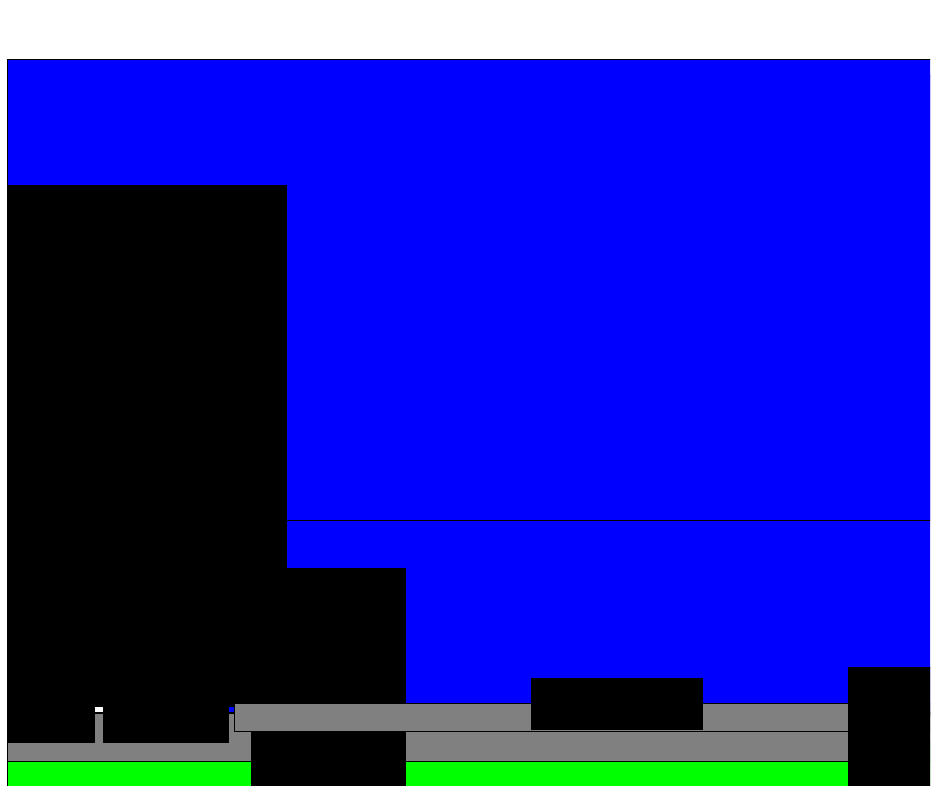}\\
& & & Black small tall at bottom left. & Black medium tall at bottom left. \\
& & & White large tall at top middle. & Black small squared at bottom left. \\
& & & Black medium tall at bottom left. & Blue large squared at center middle. \\
& & & Blue large tall at top middle. & Green medium wide at bottom middle.\\
& & & Black large tall at center left. & Grey medium wide at bottom middle. \\
\cmidrule(lr){3-5}
\end{tabular} }
\caption{Examples of block illustrations and abstract statements
for three scene images, automatically generated from (a) selective
search things proposals, and (b) human annotated things. The color
is calculated from the dominant color in the bounding box window.
Note that the human annotated objects are not complete, however,
they are a reasonable and best available approximation. }
\label{fig:AutomaticBlocksStats}
\end{figure*}

\begin{figure*}[th!]
\centering
 \scalebox{0.71}{
\begin{tabular}{c@{}c@{}c@{}c@{}}
 & \textbf{\large Human 1} & \textbf{\large Human 2} & \textbf{\large Human 3}\\
\cmidrule(lr){2-4}
 \rotatebox{90}{\textbf{\large Football field}} &
\includegraphics[width=0.33\linewidth, height=3cm, trim = 3cm 2cm 2cm 2cm, clip=true]{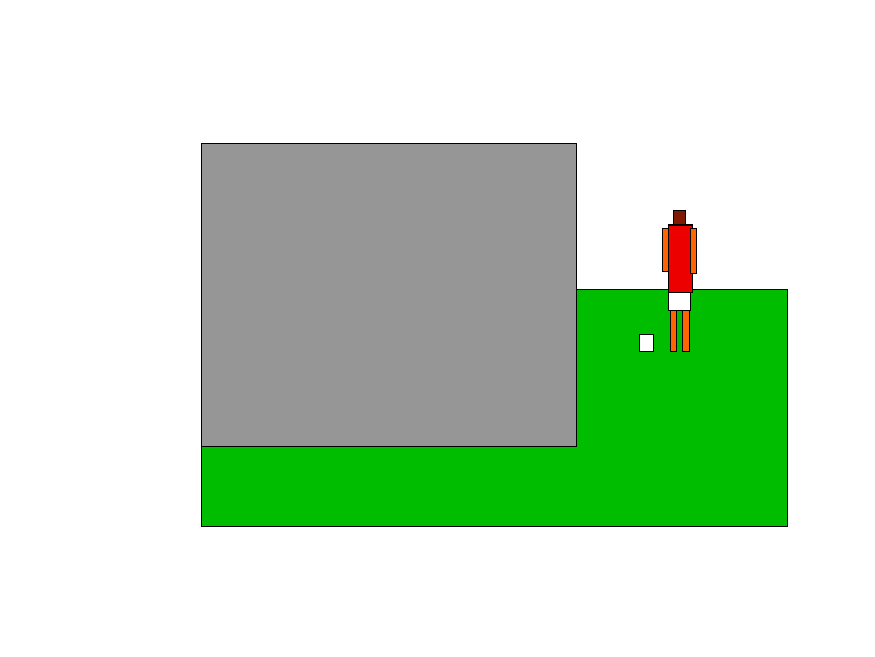}
&
\includegraphics[width=0.33\linewidth, height=3cm, trim = 3cm 2cm 3cm 0.8cm, clip=true]{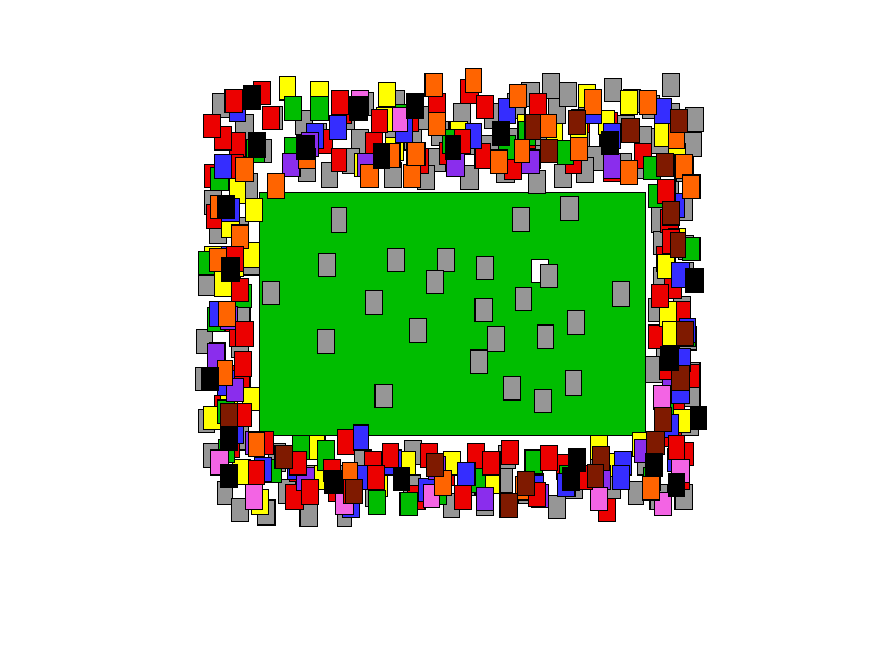}
&
\includegraphics[width=0.33\linewidth, height=3cm, trim = 3cm 1cm 3cm 1cm, clip=true]{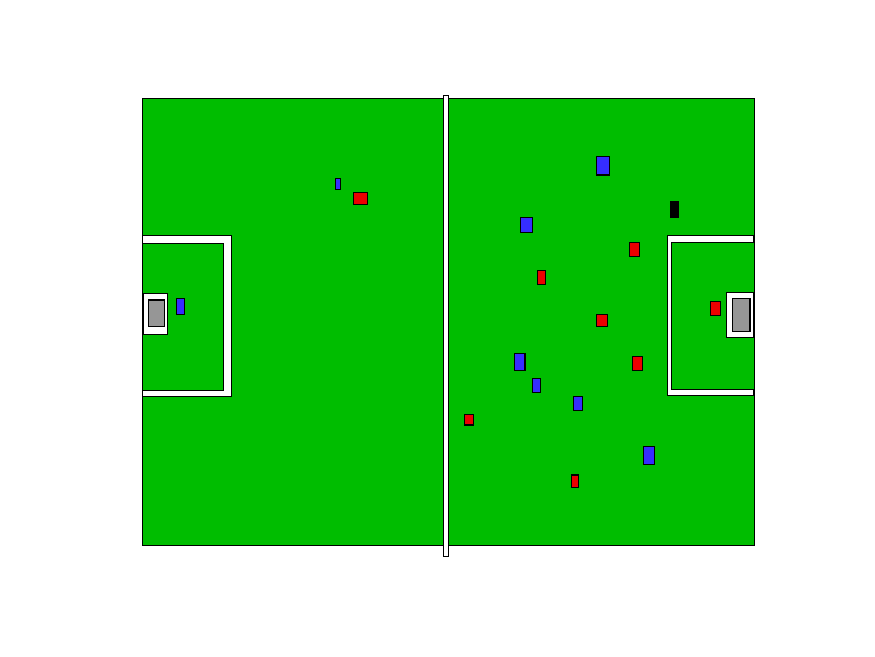}\\
 & Green large wide at bottom middle & Green large wide at center middle. & Green large wide at center middle. \\
 & White small squared at bottom middle & 'Any' small squared at top left. & White medium wide at center left. \\
 & 'Any' medium tall at center left & 'Any' small squared at bottom left. & White medium wide at center right. \\
 & White large tall at center right & 'Any' small squared at bottom right. & Red small squared at center left.\\
 &                                  & 'Any' small squared at top right. & Black small squared at bottom left. \\

\cmidrule(lr){2-4}

\rotatebox{90}{\textbf{\large Beach}} &
\includegraphics[width=0.33\linewidth, height=3cm, trim = 3cm 1.6cm 2.5cm 2cm, clip=true]{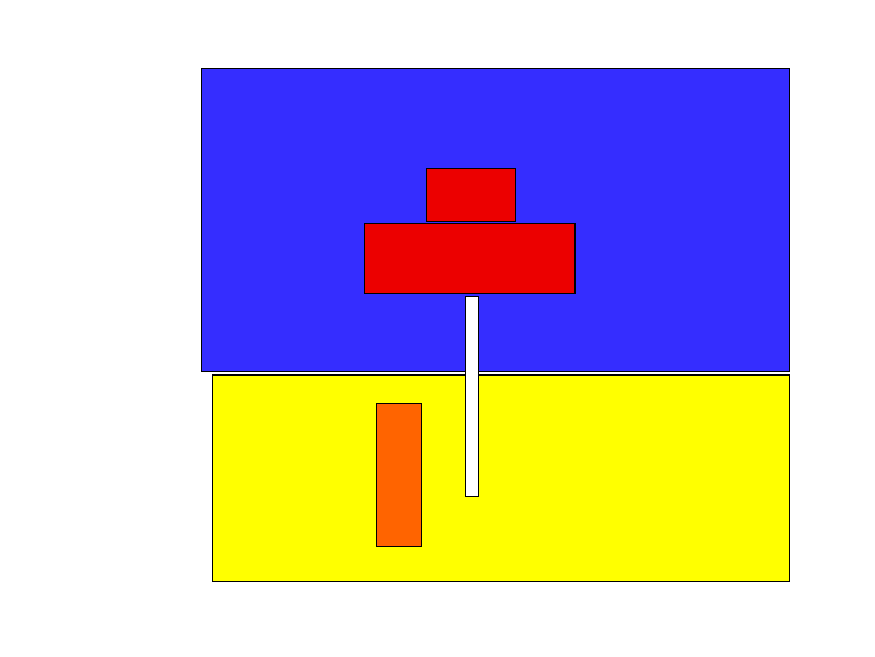}
&
\includegraphics[width=0.33\linewidth, height=3cm, trim = 3cm 1.2cm 3cm 0.8cm, clip=true]{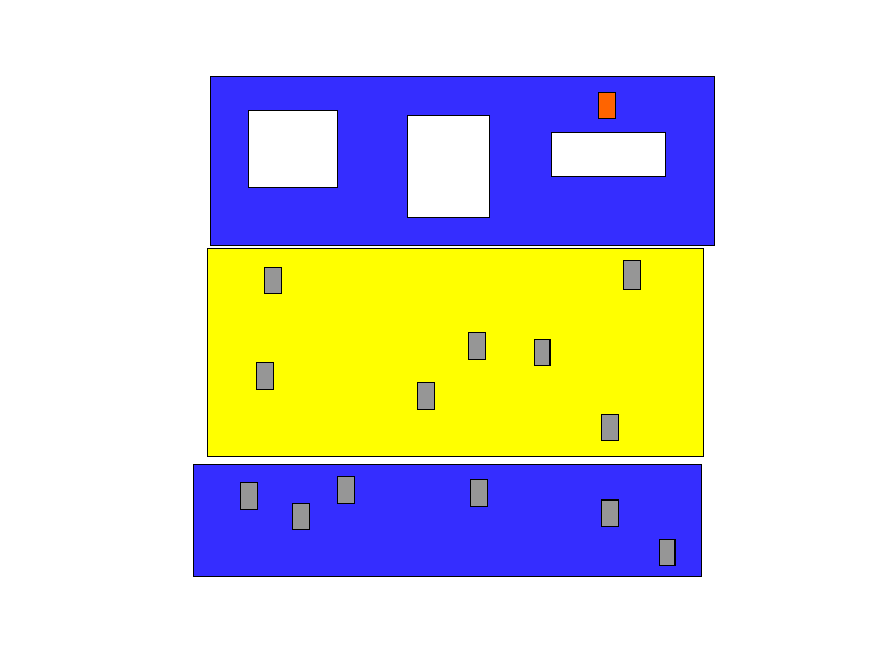}
&
\includegraphics[width=0.33\linewidth, height=3cm, trim = 3cm 1.2cm 3cm 0.8cm, clip=true]{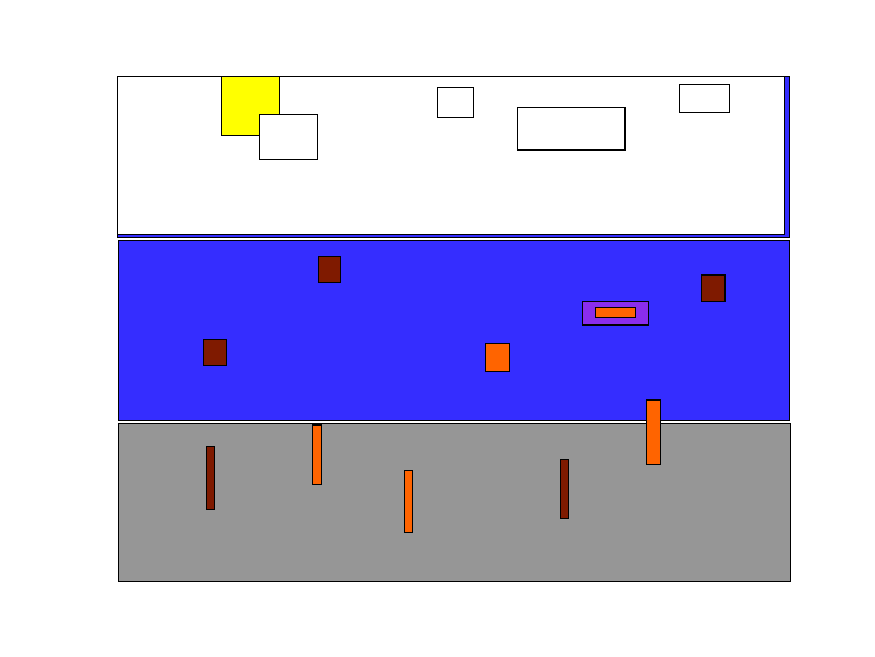}\\
 & Yellow large wide at bottom middle & Yellow large wide at bottom middle. & Blue large wide at top middle. \\
 & White medium tall at bottom left & Blue large wide at center middle. & Purple small squared at bottom right. \\
 & Blue large wide at top middle & Blue large wide at top middle. & Brown large wide at bottom middle. \\
 & Red medium wide at center left & Black medium tall at bottom right. & Yellow small squared at top left.\\
 &                                &                                    & White small squared at top middle. \\

\cmidrule(lr){2-4}

\rotatebox{90}{\textbf{\large Skyscrapers}} &
\includegraphics[width=0.33\linewidth, height=3cm, trim = 3cm 1.3cm 2cm 2cm, clip=true]{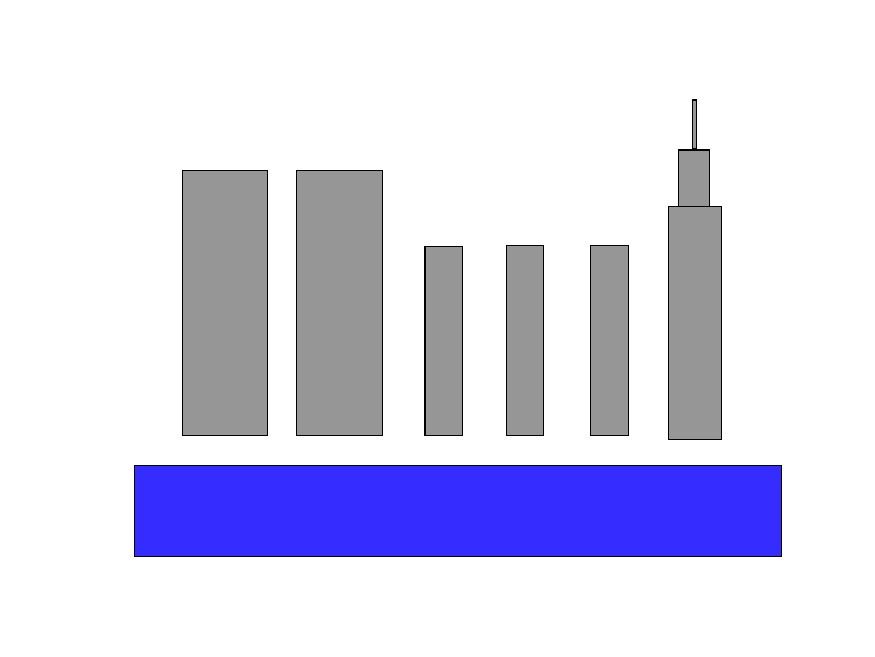}
&
\includegraphics[width=0.33\linewidth, height=3cm, trim = 3cm 1.2cm 3cm 0.8cm, clip=true]{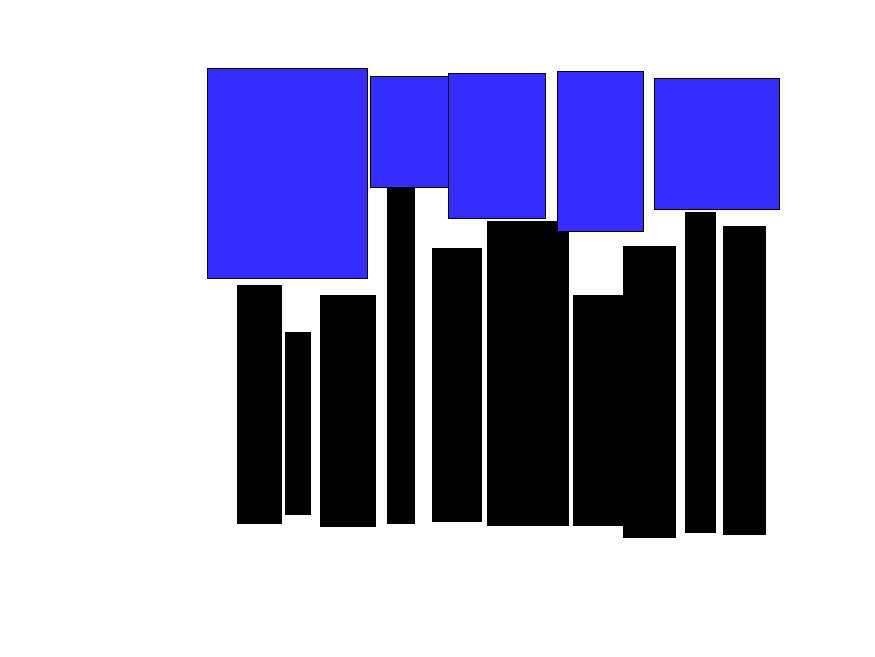}
&
\includegraphics[width=0.33\linewidth, height=3cm, trim = 1cm 1cm 1cm 0cm, clip=true]{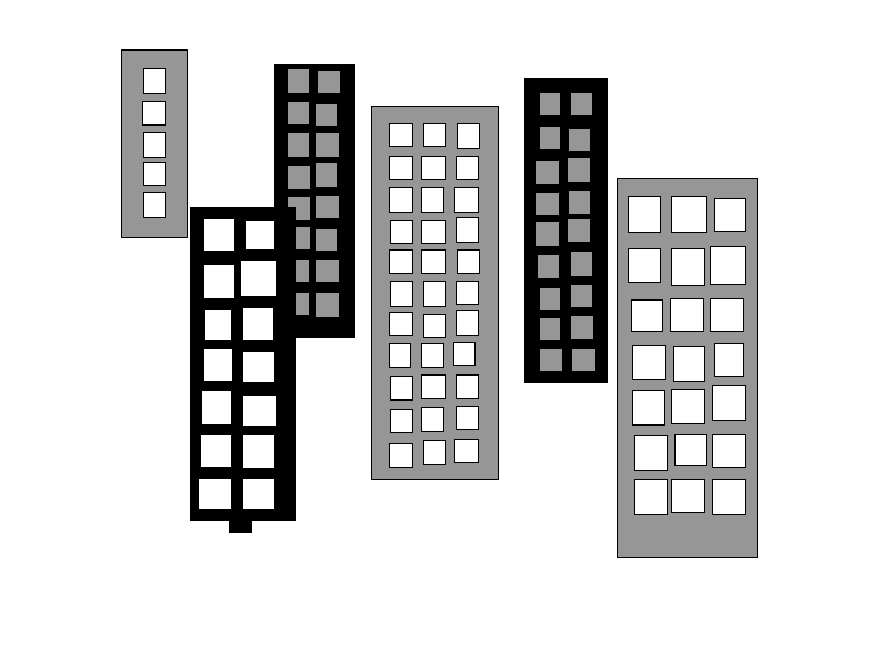}\\
 & Grey large tall at center middle & Blue large wide at top middle. & Grey large tall at bottom middle. \\
 & Grey large tall at top middle & Black medium tall at bottom left. & Grey large tall at bottom left. \\
 & Grey large wide at bottom middle & Black medium tall at bottom left. & Grey large tall at bottom right. \\
 &                                  & Black medium tall at bottom right. & Grey large tall at center left.\\
 &                                  &                                    & Grey large tall at center right. \\
\cmidrule(lr){2-4}
\end{tabular} }
\caption{Block illustrations and abstract statements manually
created. We asked three independent humans to provide statements
and block illustrations for three scenes using only their memory
of a scene, without looking at any image examples. Interestingly,
we observe that sensible descriptions can be provided in this
manner, even though this is not the way humans normally
communicate or describe scenes in everyday life. Additionally,
they are quite diverse for each individual.}
\label{fig:AbstractStatsAndBlockIll}
\end{figure*}

For the experiments we use both standard scene datasets, and
abstract datasets we automatically generate from object
annotations. Creating abstract datasets in this manner, examples
shown in Figure~\ref{fig:AutomaticBlocksStats}, is different from
when statements and block illustrations are provided directly by
humans, see Figure~\ref{fig:AbstractStatsAndBlockIll} for
comparison. However, this is the best approximation we can get for
free. We describe the datasets below.

\textbf{SUN2012-14Scenes} has 1,400 images in 14 classes we select
from SUN397~\cite{xiaoCVPR10sunDB}, with an objective to have
object annotations for at least 100 images per scene. We use this
dataset for comparison between properties of human annotated
things and properties of automatically generated things by
~\cite{alexe2010object, uijlingsIJCV13selectiveSearch,
manenICCV13objectProposals}.

\textbf{Indoor67}~\cite{quattoniCVPR09indoorSceneDB} has 15,620
images from 67 indoor scenes like \emph{bedroom, restaurant,
winecellar}. All scenes have at least 100 images per category,
with a provided 80-20 training-test splits. For experiment 1, we
use the training-test split as a given. For experiment 2 and 3,
since we do not need any examples for training, we use the 80
images as test images for searching without examples. The
remaining 20 images we use as a holdout set, where we ignore their
class annotations.

\textbf{SUNAttributes}~\cite{Patterson2012SunAttributes} contains
14,340 images hierarchically grouped in 3 levels, starting from
fine-grained scenes in level three, growing into more general
scene categories in level two and one. Level one has 3 general
classes: \textit{indoor}, \textit{outdoor natural} and
\textit{outdoor man made}. Level two has 16 high level scene
categories like \emph{shopping and dining} or \emph{water, ice,
snow}, and level three has all 717 fine-grained scenes like
\emph{airport ticket counter, bicycle racks, canyon}, with 20
images for each scene, which we use for testing.

\textbf{Indoor67-AbstractStatements}. Many of the Indoor67 images
have object annotations provided with the LabelMe toolbox
\cite{LabelMeRussell2008} in a shape of a polygon. We ignore the
object names, consider them as nameless things and calculate their
properties from a bounding box surrounding the polygon. As color
we use the most dominant color in the bounding box, computed as in
\cite{colornaming2009}. From the things properties we generate
sentences for free as described in Section~\ref{sec:statements},
and use them to generate 67 scene representations, in a form of a
histogram, for each scene to query by.

\textbf{Indoor67-AbstractBlocks}. Similarly as in the
AbstractStatements-67 dataset, we reuse the LabelMe annotations
from Indoor67, to automatically generate abstract block
illustrations. We ignore the objects type and appearance, and we
use the position, size, ratio and the dominant color of the things
properties to generate block illustrations. From the block
illustrations we compute Fisher vector representations for each of
the 67 scenes to query by.

\textbf{SUN717-AbstractStatements}. The SUNAttributes dataset also
comes with object annotations from the LabelMe toolbox in the form
of polygons. Similar as in the creation of
Indoor67-AbstractStatements, we ignore the object annotations and
generate bounding boxes around the polygons to hold things. For
all image things we generate abstract statements. By grouping the
abstract statements per image of all three levels from
SUNAttributes, we generate 3, 16 and 717 scene representations of
abstract statements respectively to query by.

\textbf{SUN717-AbstractBlocks}. We use the thing properties
calculated from the object annotations on SUNAttributes to also
generate abstract block illustrations. From the block
illustrations we create 3, 16 and 717 scene class representations,
in a Fisher vector form, for level one, two and three
respectively.

We will make all abstract datasest available online upon
acceptance.

\section{Experiments}
\subsection{Comparing automatically proposed and manually annotated things} \label{exp:VersusHuman}

\setlength{\tabcolsep}{4pt}
\begin{table}[t!]
\centering
\scalebox{0.82}{
\begin{tabular}{lccccc} \toprule
     & \textbf{Horizontal} & \textbf{Vertical} & \textbf{Size} &
\textbf{Ratio} & \textbf{Color}\\ \midrule
Objectness~\cite{alexe2012measuring}                  & 0.073           & 0.029            & \textbf{0.200}     & 0.179            & 0.286\\
Selective search~\cite{uijlingsIJCV13selectiveSearch} & \textbf{0.043}  & 0.043            & 0.360              & 0.066            & \textbf{0.282}\\
PRIMObjects~\cite{manenICCV13objectProposals}        & 0.058           & \textbf{0.028}   & 0.230              & \textbf{0.061}   & 0.294\\
\bottomrule
\end{tabular}
} \caption{ Comparison between things properties distribution of
three things proposal methods and human annotated things on the
SUN2012-14Scenes dataset, measured with the Kullback-Leibler
divergence. The results are computed by averaging the KL
divergence over all scene classes. The closest distributions are
shown in bold. Each proposal method has a best approximation to
the human annotated things for different property, each successful
in their own right. When we compare this KL values with the
maximum inter-class KL divergences of human things properties as
reference numbers, 0.584 for horizontal, 0.248 for vertical, 0.651
for size, 0.292 for ratio and 6.068 for color, as shown in
Figure~\ref{fig:HeatMaps_SS_Humans}, we conclude that the
properties of things proposal methods come close to the properties
of human annotated things. } \label{table:KL-div}
\end{table}
\setlength{\tabcolsep}{1.4pt}

\begin{figure*}[th!]
\centering
\includegraphics[width=\linewidth, trim=0in 0in 0in 0in, clip=true]{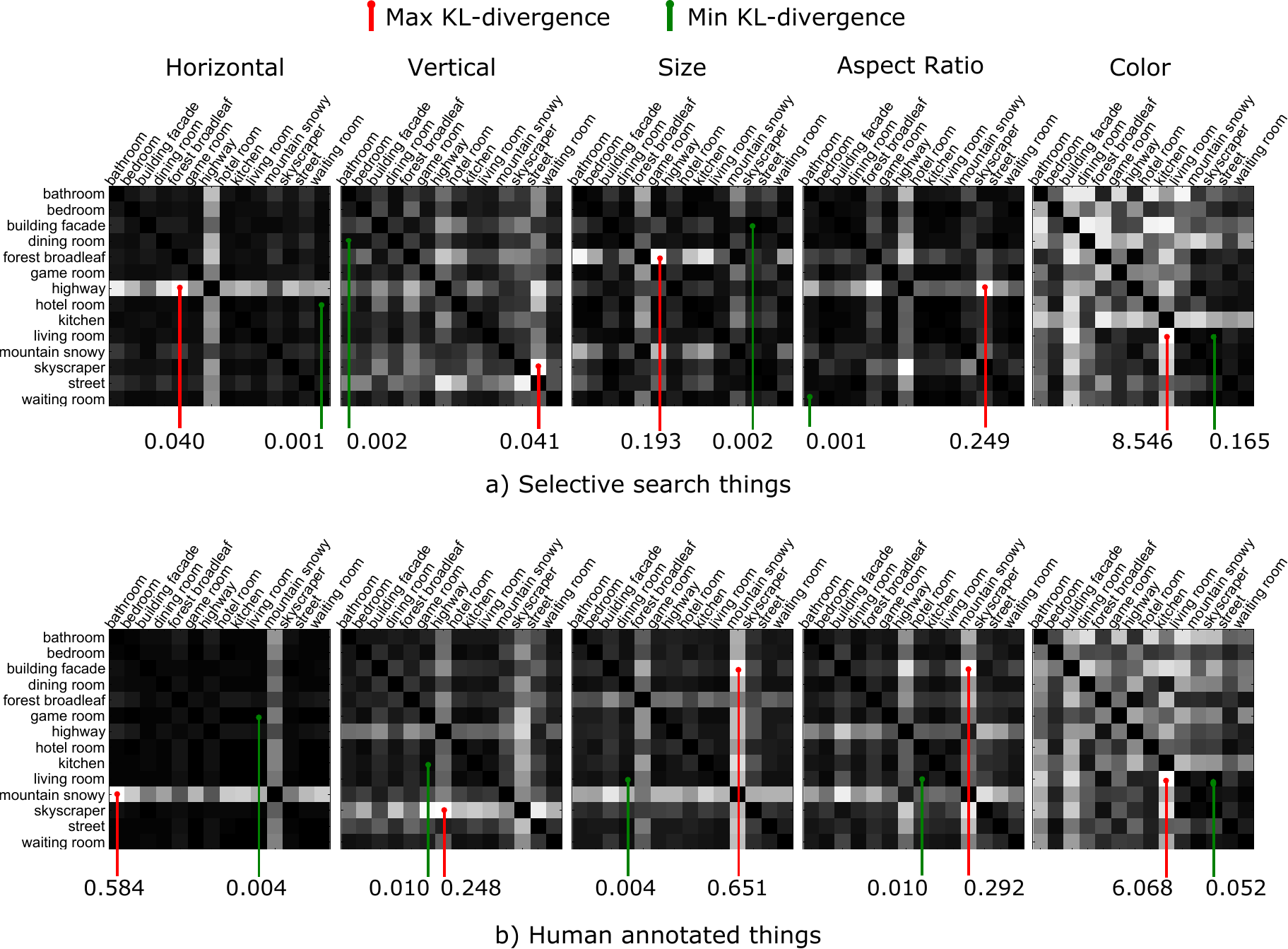}
\caption{ Heat maps per property of Kullback-Leibler (KL)
divergence between scenes, for (a) selective search things and (b)
human annotated things, with pointed maximum (red) and minimum
(green) values. The lower the KL divergence score, the more
similar the scenes are for that property. Interestingly, we see a
similar pattern between the properties of selective search things
and human annotated things, indicating the suitability of things
proposal methods to approximate true things as annotated by
humans.} \label{fig:HeatMaps_SS_Humans}
\end{figure*}


\textbf{Do they behave similar?} We first consider the question
whether properties from things obtained with proposal methods are
distributed similarly as things taken from manual things
annotations. We generate things proposals with three recent
methods: objectness~\cite{alexe2012measuring}, selective
search~\cite{uijlingsIJCV13selectiveSearch} and prime object
proposals (PRIM)~\cite{manenICCV13objectProposals}, and we
consider manually annotated things from the SUN2012-14Scenes
dataset. Although the human annotations in SUN2012-14Scenes are
not intended to be a complete ground truth of all things contained
in a scene, they are the best approximation available to us. We
compute the distributions of things properties per image by
binning. We measure the distribution divergence of things
proposals property $Q$ to the human annotated thing property $P$
with the Kullback-Leibler divergence $D_{KL}(P||Q)=\sum_i \ln
\left( \frac{P(i)}{Q(i)} \right) P(i)$. The Kullback-Leibler
divergence measures the expected loss of information when
distribution $Q$ is used to approximate the true distribution $P$.
A low score represents a good fit.

\begin{figure*}[t!]
\centering
\includegraphics[width=\linewidth, trim=1.1in 3.8in 1.2in 1in, clip=true]{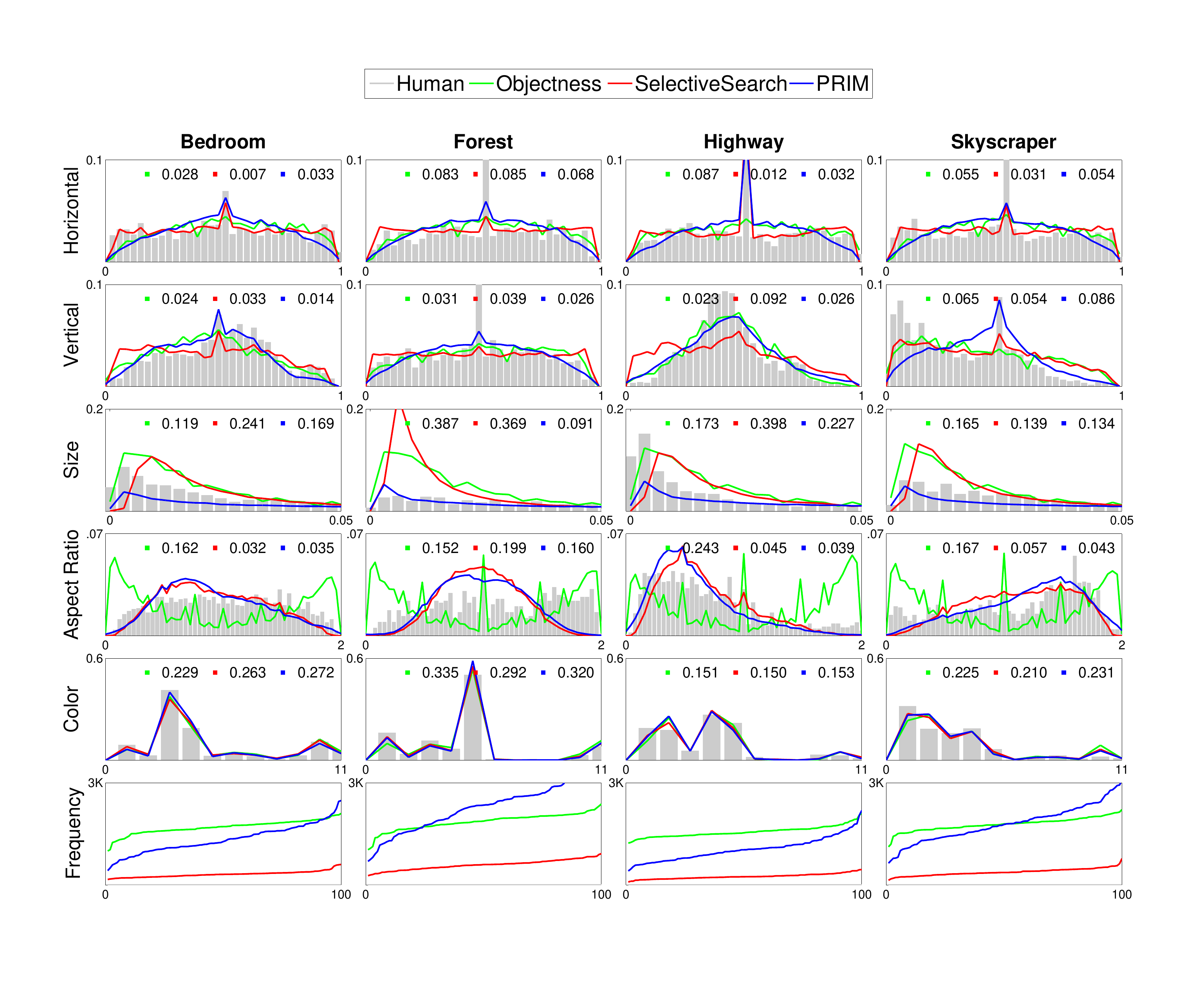}
\caption{ Comparing properties of human annotations and things
proposals. The gray bars represent distributions of properties
calculated from human annotated things, and colored lines show the
distribution of properties calculated from things proposals. The
numbers over the plot lines are the KL-divergence of human vs
proposed things. In most cases the thing proposal distribution
lines follow the shape of the gray bars of human annotated things.
Thus, we conclude that the things proposals are a suitable
approximation of real things in a scene.}
\label{fig:DistributionFits}
\end{figure*}

\textbf{Analysis.} We present the mean Kullback-Leibler divergence
scores over all scenes, between the things properties of proposal
methods and human annotated things in~Table~\ref{table:KL-div}.
The lower the KL divergence score, the more similar the
distributions are for that property. Results show that proposals
reasonably approximate human annotation statistics with an
expected average loss of only 5-10\%, except for \textit{size} and
\textit{color}. The difficulty in estimating the color is due to
its computation over the full window, \ie, it is a joint statistic
over all four dimensions and thus affected by errors in any of
them. If we take as reference numbers the maximum inter-class KL
divergences of human things properties, with 0.584 for horizontal,
0.248 for vertical, 0.651 for size, 0.292 for ratio and 6.068 for
color, as shown in Figure~\ref{fig:HeatMaps_SS_Humans}, we
conclude that the properties of things proposal methods come close
to the properties of human annotated things. In
Figure~\ref{fig:HeatMaps_SS_Humans} we also show the minimum
values and heat maps of the KL divergence between scene classes
per property, for both human annotated things and selective search
things. Interestingly, we see a similar pattern between the
properties of selective search things and human annotated things.
For color, the same scenes have max/min values. For horizontal
position of selective search things, \emph{highway} differs most
from the rest of the scenes. We believe this is so because a
highway scene usually has things lined horizontally, like the
highway, sky, ground, which is not the case within the other 13
scenes. However, we observed that the human annotated things of
highway are mostly on the cars on the highway, missing the
horizontally aligned things in the scene. Interestingly, some of
the most similar scenes are, for horizontal: \emph{hotel room -
waiting room}, size: \emph{building facade - skyscraper} and
aspect ratio: \emph{waiting room - bathroom}. Most distinctive
are, for horizontal: \emph{highway - forest broadleaf}, vertical:
\emph{skyscraper - street} and aspect ratio: \emph{highway -
skyscraper}. The examples show that the things properties capture
scene information, the similar scenes are indeed close, and the
dissimilar ones are far for a given things property. Investigating
the matter further, we show the distribution curves for the
properties of human annotations and all three things proposals for
four scenes in Figure~\ref{fig:DistributionFits}. All proposed
things follow the distributions of the horizontal and vertical
position properties close to the horizontal and vertical position
properties of human annotated things. Object size is
overestimated, which may be due to the proposals used in object
recognition where larger objects are the norm. Aspect-ratio is
coming close reasonably well, except objectness, which tends to
generate overly long or broad windows. The things proposals are
not perfect, but we conclude that they are a suitable
approximation of real objects in a scene. As new and better object
proposal methods are being introduced \cite{Dollar2015PAMI,
RenHG015}, we can expect that the approximation will only improve.

\textbf{Are they as discriminative?} Next we study whether the
object proposal window properties are as discriminative as human
annotated things. We use the SUN2012-14Scenes dataset with 50
images for training and 50 for testing, and we use the Indoor67
dataset with 67 indoor scene classes using the author suggested
splits. As representation we rely on the things syntax with the
Fisher vector, calculated using a 1,024 component GMMs. For
learning we train one-vs-rest Support Vector Machine with the RBF
kernel, and we evaluate with accuracy.

\textbf{Results}. When using as representation the things syntax
of human annotations encoded with the Fisher vector, on
SUN2012-14Scenes we achieve an accuracy of 57.6\%. The Fisher
vectors of things syntax from automatically generated things
proposals with selective search come close with an accuracy of
54.0\%, with PRIM proposals 52.1\%, and 32.2\% for objectness
things. For comparison, On 14 classes the random recognition rate
is 7.14\%. The results show that the things syntax has a
discriminative potential, and the discriminative potential of
things proposals, especially the one from selective search,
approximates the discriminative potential of human annotated
things reasonably. Thus, we use things proposals from selective
search in the rest of our experiments.

On Indoor67 the things syntax of selective search encoded with the
Fisher Vector achieves an accuracy of 25.8\%. The accuracy with
GIST \cite{olivaIJCV01spatialEnvelope} is 29.6\%, and when GIST is
combined with the things syntax the accuracy reaches 38.9\%. This
shows that the things syntax is orthogonal to GIST, capturing new
information. We also recognize the advances in deep learning and
the power of the features of a Convolutional Neural Network (CNN)
\cite{KrizhevskySH12, DeepNetEccvZeilerF14}. For example, when
using features from one layer before the last of GoogleNet
\cite{SzegedyCVPR2015}, the accuracy on Indoor67 is 66.8\%. When
we combine these features with the things syntax, the result
improves slightly to 68.5\%. This shows that the things syntax
also captures some new information not learned by the CNN.
Overall, we show that the things syntax has some discriminative
power. Yet, the main aim of the things syntax is not to improve
the scene classification, the CNN features do a much better job at
it. The aim and the added benefit is to generate an example-free
abstract description without the need for learning, where all the
others, including deep learning, use examples to learn human
understandable representations.

Overall, in this experiment we have shown that the things syntax
holds discriminative information, that the information from
proposed things comes close to human annotated things. We
investigate the abstract descriptions from the things syntax in
the next experiments.

\subsection{Query by abstract statements}

\begin{figure*}[t!]
\centering
\begin{tabular}{@{}c@{\hspace{5px}}c@{\hspace{5px}}c@{\hspace{5px}}c@{}}
\includegraphics[width=0.42\linewidth, height=5cm]{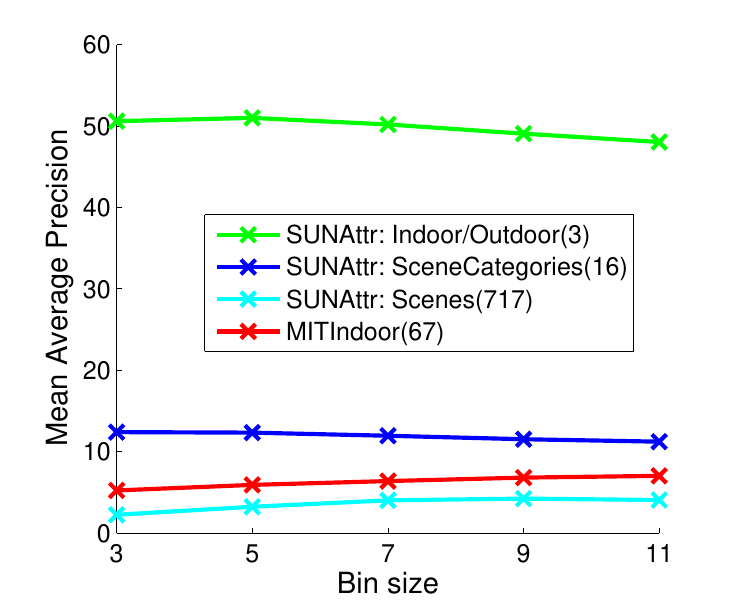} &
\includegraphics[width=0.16\linewidth, height=5cm, trim = 6mm 0mm 0mm 0mm, clip=true]{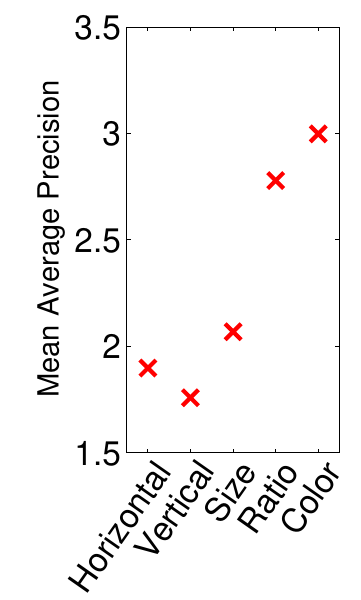} &
\includegraphics[width=0.42\linewidth, height=5cm]{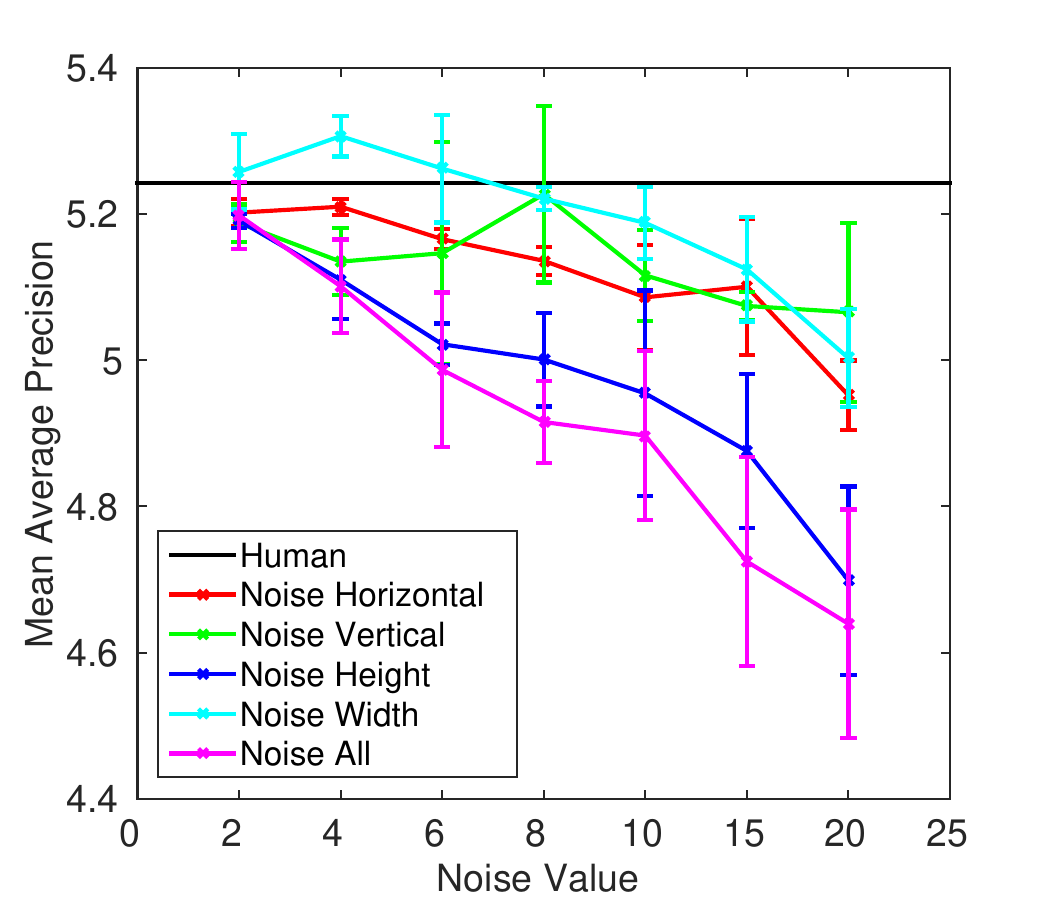} \\
(a) & (b) & (c) \\
\end{tabular}
\caption{Result of investigating abstract statements for
retrieving scenes: (a) influence of the number of bins used to
quantize the things syntax into histograms and statements, (b)
descriptive influence per property on Indoor67, (c) influence of
abstract statements quality by adding noise to object annotations
on Indoor67.} \label{fig:ResultsTell}
\end{figure*}

In this experiment we investigate the use of abstract statements
to describe a scene, like ``blue wide thing at top center, green
medium tall thing at bottom left'', as described in
Section~\ref{sec:statements}, and their descriptive potential in a
retrieval setting. We use statements from the
SUN717-AbstractStatements and Indoor67-AbstractStatements
datasets, and test on the images from SUNAttributes and Indoor67.
We investigate three parameters of the abstract statements for
scene retrieval, evaluate with mean average precision (MAP), and
we summarize the results in Figure~\ref{fig:ResultsTell}.

\textbf{Influence of binning.} First, we investigate the influence
of the number of bins for horizontal position, vertical position,
size and aspect ratio. Rather than restricting the binning to
three annotations for objects, like (\emph{left, middle, right})
or (\emph{top, center, bottom}), we can easily add more bins to
represent the statements, \eg (\emph{most-left, left, center,
right, most-right}). The number of bins for color is always fixed
to 11, as defined in \cite{colornaming2009}. We test on images of
all three levels from the SUNAttributes dataset and Indoor67. In
Figure~\ref{fig:ResultsTell} (a), we show retrieval results with
bin sizes varying from 3 to 11. We observe that on more general
categories, like level one and two of SUNAttributes, using more
bins/more precise descriptions of the things properties does not
help, whereas, for fine-grained scenes it results in better
retrieval MAP. For example on the SUNAttributes 717 scenes the
mean average precision grows from 2.25\% to 4.06\%, and on
Indoor67 from 5.24\% to 7.04\%. We conclude, when more precise
statements are available for fine-grained scenes, the better the
scene retrieval will be.

\textbf{Influence per property.} We investigate how well the thing
properties perform independently. In this way, we generate a scene
representation for each property, by having statements composed of
one word only, as ``left'' things, or ``tall'' things, or
``small'' things. We create scene representations from
AbstractStatements-67 and evaluate on Indoor67. The results are
shown in Figure~\ref{fig:ResultsTell} (b). Ratio and color are the
best performing properties. We assume this is because they are
both scale invariant, whereas the other properties are not. For
example, if things are captured closely or from far away, the
aspect ratio and color will be consistent, whereas the position
and size of things in the image will change.

\textbf{Influence of statements quality}. The abstract statements
we generate in the AbstractStatements-67 and the
SUN717-AbstractStatements datasets are by using human annotated
things. To approximate a more realistic scenario where the
abstract statements will be provided directly from users, we add
noise to the bounding box things annotations. We generate the
noise from a Gaussian distribution with mean 0, and standard
deviations of [2, 4, 6, 8, 10, 15, 20] pixels, and we scale the
images to a maximum dimension of 320px. We investigate adding
noise on Indoor67-AbstractStatements and test on Indoor67. We add
noise on each property separately, as well as on all properties
together and present results in Figure~\ref{fig:ResultsTell} (c).
As expected, adding noise has some effects the retrieval results.
For example when adding noise on all thing properties together up
to 10px, the MAP goes down from 5.24\% to 4.90\%, and for 20px
noise it goes down to 4.64\%. Interestingly adding noise up to 6px
on the things width, the results even improve marginally. This
happens when a mistake is made with the binning, the noise acts as
a correction mechanism. Overall, we conclude that the quality of
the object location can tolerate displacement without hurting the
scene retrieval substantivally.

\subsection{Query by block illustrations}

\begin{figure*}[t!]
\centering
\begin{tabular}{@{}c@{\hspace{5px}}c@{\hspace{5px}}c@{\hspace{5px}}c@{}}
\includegraphics[width=0.42\linewidth, height=5cm]{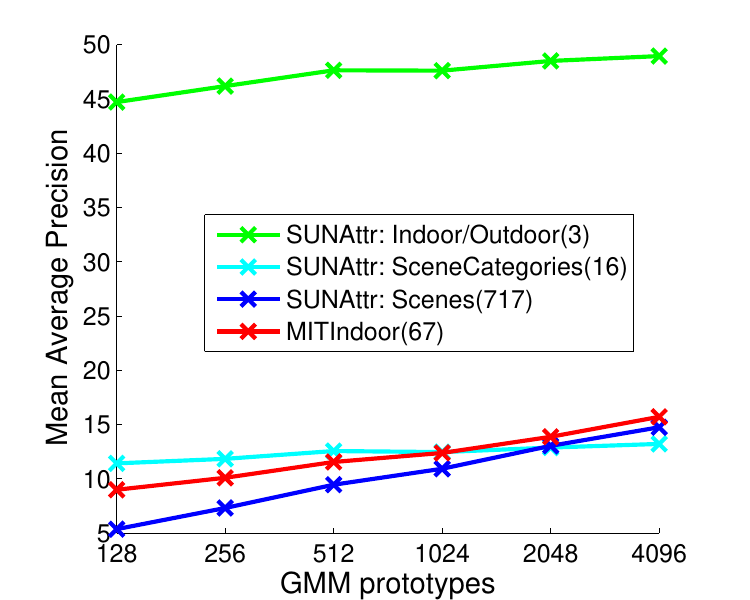} &
\includegraphics[width=0.16\linewidth, height=5cm, trim = 6mm 0mm 0mm 0mm, clip=true]{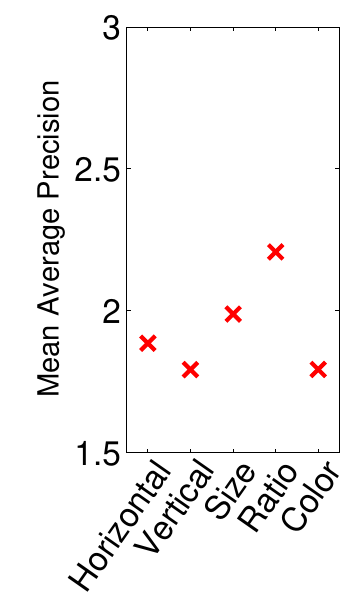} &
\includegraphics[width=0.42\linewidth, height=5cm]{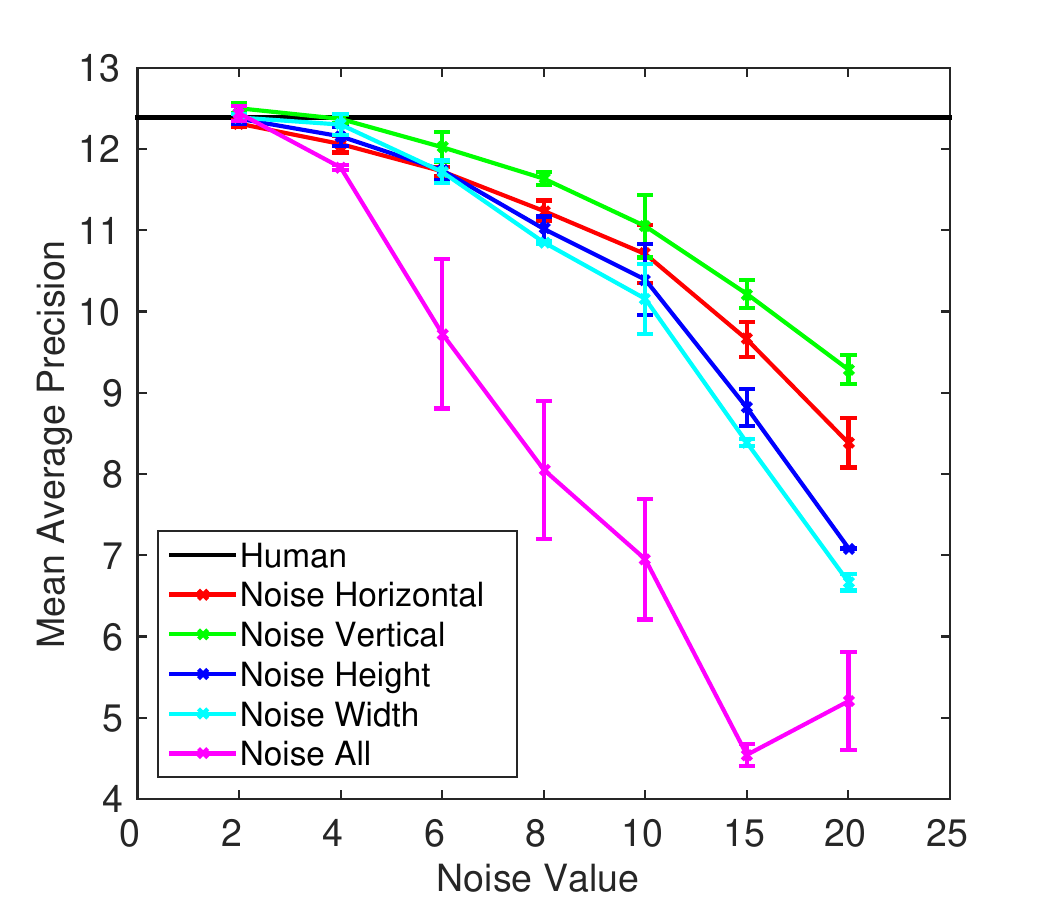} \\
(a) & (b) & (c) \\
\end{tabular}
\caption{Result of investigating block illustrations for
retrieving scenes: (a) influence of the number of Gaussian Mixture
Model (GMM) prototypes used in encoding the block illustrations
with the Fisher vector, (b) descriptive influence per property on
Indoor67, (c) influence of Fisher vector representations for block
illustrations generated from human annotations, by adding noise to
the annotations on Indoor67.} \label{fig:ResultsDraw}
\end{figure*}

In the third experiment we investigate to what extent we can
retrieve an unseen scene using block illustrations of things. As a
representation of block illustrations and things syntax from test
images we use Fisher vector encoding, as described in
Section~\ref{sec:blockillustrat}. We use Fisher vector
representations per scene, created from block illustrations of
SUN717-AbstractBlocks and Indoor67-AbstractBlocks datasets, to
query scenes in test images from SUNAttributes and Indoor67. We
investigate three parameters of the Fisher representation in a
scene retrieval setting, we evaluate with mean average precision,
and summarize the results in Figure~\ref{fig:ResultsDraw}.

\textbf{Influence of GMM components}. The Fisher vector encoding
depends on the number of GMM prototypes, therefore, we investigate
their influence. We compute the GMM prototypes on a holdout set of
Indoor67. In Figure~\ref{fig:ResultsDraw} (a) we show results on
SUNAttributes, including all three levels, and Indoor67. As
expected, since more prototypes capture a better variation of the
windows properties, they result in richer Fisher vector
representation and higher MAP. On the third level of SUNAttributes
the MAP grows from 5.37\% for 128 prototypes, to 14.75\% for 4,069
prototypes. On Indoor67 similar improvement fashion of the MAP is
followed, from 8.99\% to 15.71\%. The results grow slowly after
1,024 prototypes. Therefore we use 1,024 component GMMs in the
rest of the experiments.

\textbf{Influence per property.} We investigate scene retrieval
using query by block illustrations per property. To do so, we
generate scene Fisher representations from only one window
property at a time from the Indoor67-AbstractBlocks dataset and
test on the Indoor67 dataset. We show the results in
Figure~\ref{fig:ResultsDraw} (b). The MAP results follow similar
trend as in the scene retrieval setting with abstract statements,
with ratio being most descriptive. Color shows a different
behavior, having a lower MAP from size and ratio. We believe this
is so, because in the histogram representations of statements we
use exactly 11 bin values for color, thus being more precise,
whereas the Fisher vector encodes the color with 1024 GMMs.

\textbf{Influence of block illustrations quality.} In the
Indoor67-AbstractBlocks dataset, we generate the block
illustrations from human annotated things. In order to approximate
a more realistic drawing scenario, we add Gaussian noise to the
human annotations and summarize the results in
Figure~\ref{fig:ResultsDraw} (c). If we add noise only to the
position, or only to the width or height of the windows, the MAP
is moderately affected up to a deviation of 8 pixels. If noise is
added to all the window properties at once the MAP drops more
rapidly. We conclude that abstract block drawings can be used to
retrieve scenes when the boxes in the block illustrations deviate
up to around 8 pixels from the true objects.

\setlength{\tabcolsep}{4pt}
\begin{table*}[t!]
\centering \scalebox{0.75}{
\begin{tabular}{lcccc}
\hline
 & \textbf{MIT Indoor} & \multicolumn{3}{c}{\textbf{SUN Attributes}} \\
 \cmidrule(lr){2-2} \cmidrule(lr){3-5}
 & \textit{Scenes(67)} & \textit{Indoor/Outdoor(3)} & \textit{SceneCategories(16)} & \textit{Scenes(717)}\\
\hline
Query by object attributes                             &  2.06\% & 34.85\% &  6.66\% &  0.33\% \\
Query by object bank                                   &  1.96\% & 45.92\% &  8.95\% &  0.37\% \\
Query by classemes                                     &  2.96\% & \textbf{60.13\%} & 13.27\% &  0.82\% \\
Query by 1,000 objects CNN-layer                       & 10.15\% & 57.81\% & 15.43\% &  2.09\% \\
\cmidrule(lr){1-1}
Query by abstract statements                           & 5.25\%  & 50.62\% & 12.40\% &  2.25\% \\
Query by block illustrations                           & 12.39\% & 47.66\% & 12.49\% &  \textbf{10.94\%} \\
Query by statements and blocks                         & \textbf{12.62\%} & 56.67\% & \textbf{17.10\%} &  9.64\% \\
\hline
\end{tabular}
} \caption{Comparison of scene retrieval MAP results, using query
by semantic representations and query by our abstract
representations, statements and block illustrations. Surprisingly,
by using abstract representations of simple things properties and
no learning at all, we manage to retrieve scenes reasonably. The
abstract representations in some cases show even better retrieval
performance than the semantic representations, which require
learning beforehand.} \label{table:ZeroShotResults}
\end{table*}
\setlength{\tabcolsep}{1.4pt}

\subsection{Abstract vs Semantic}

\begin{figure*}[t!]
\centering
\begin{flushright}
\scalebox{1}{
\begin{tabular}{@{}c@{}c@{}c@{}}
Abstract statements & Block illustrations\\
\includegraphics[clip=true, trim=0in 0in 0in 0in, width=0.5\linewidth]{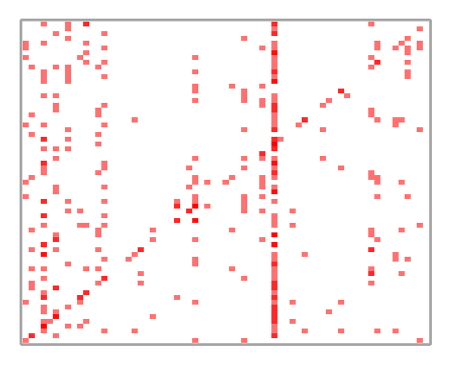} &
\includegraphics[clip=true, trim=0in 0in 0in 0in, width=0.5\linewidth]{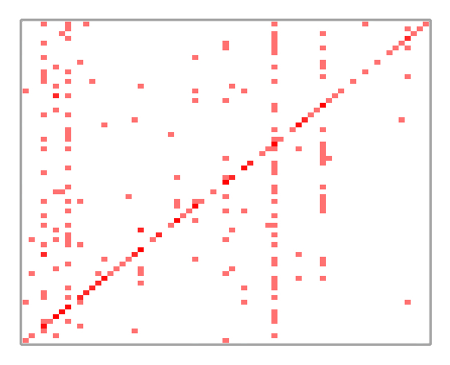}\\
\end{tabular}}
\end{flushright}
\begin{flushright}
\scalebox{1}{
\begin{tabular}{@{}c@{}c@{}c@{}c@{}}
Object attributes & Object bank & Classemes & 1,000 ImageNet \\
\includegraphics[clip=true, width=0.25\linewidth]{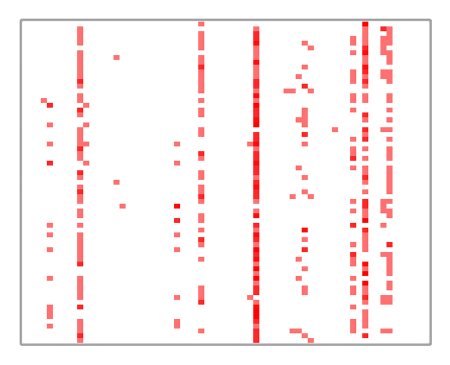} &
\includegraphics[clip=true, width=0.25\linewidth]{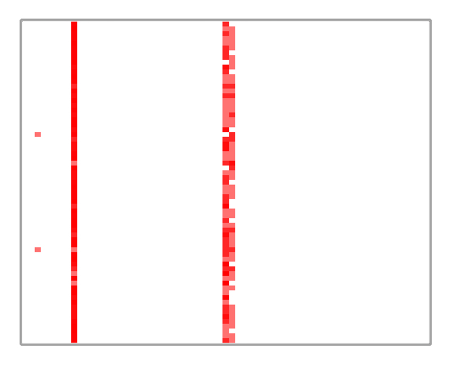} &
\includegraphics[clip=true, width=0.25\linewidth]{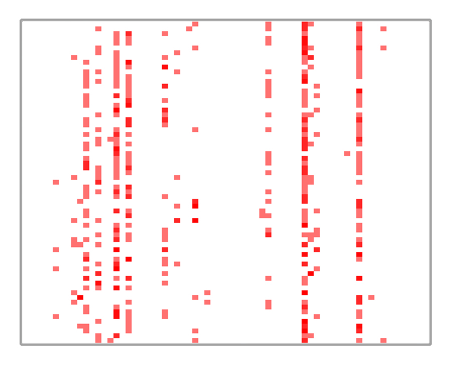} &
\includegraphics[clip=true, width=0.25\linewidth]{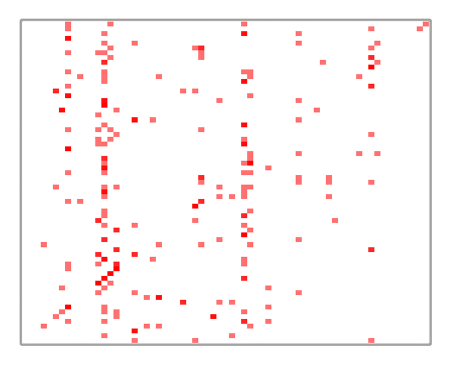}
\\
\end{tabular} }
\end{flushright} \caption{Confusion matrices on the Indoor67 dataset of our abstract
representations and other semantic representations. In the
confusion matrixes of the abstract representations, the diagonal
line can be clearly seen, depicting true positives for all scene
classes. In the semantic representations most scenes get
misclassified into few scenes that have an accidental overlap with
a semantic class, which can be seen with the strong vertical red
line responses in the confusion matrices. Overall, when there is
no direct semantic information available aimed for the scope of
the unseen classes, abstract descriptions of things syntax are a
good alternative representation. }
\label{fig:ConfussionMatricesAbstVsSens}
\end{figure*}

\begin{figure*}[tbp]
\centering
\includegraphics[width=\linewidth]{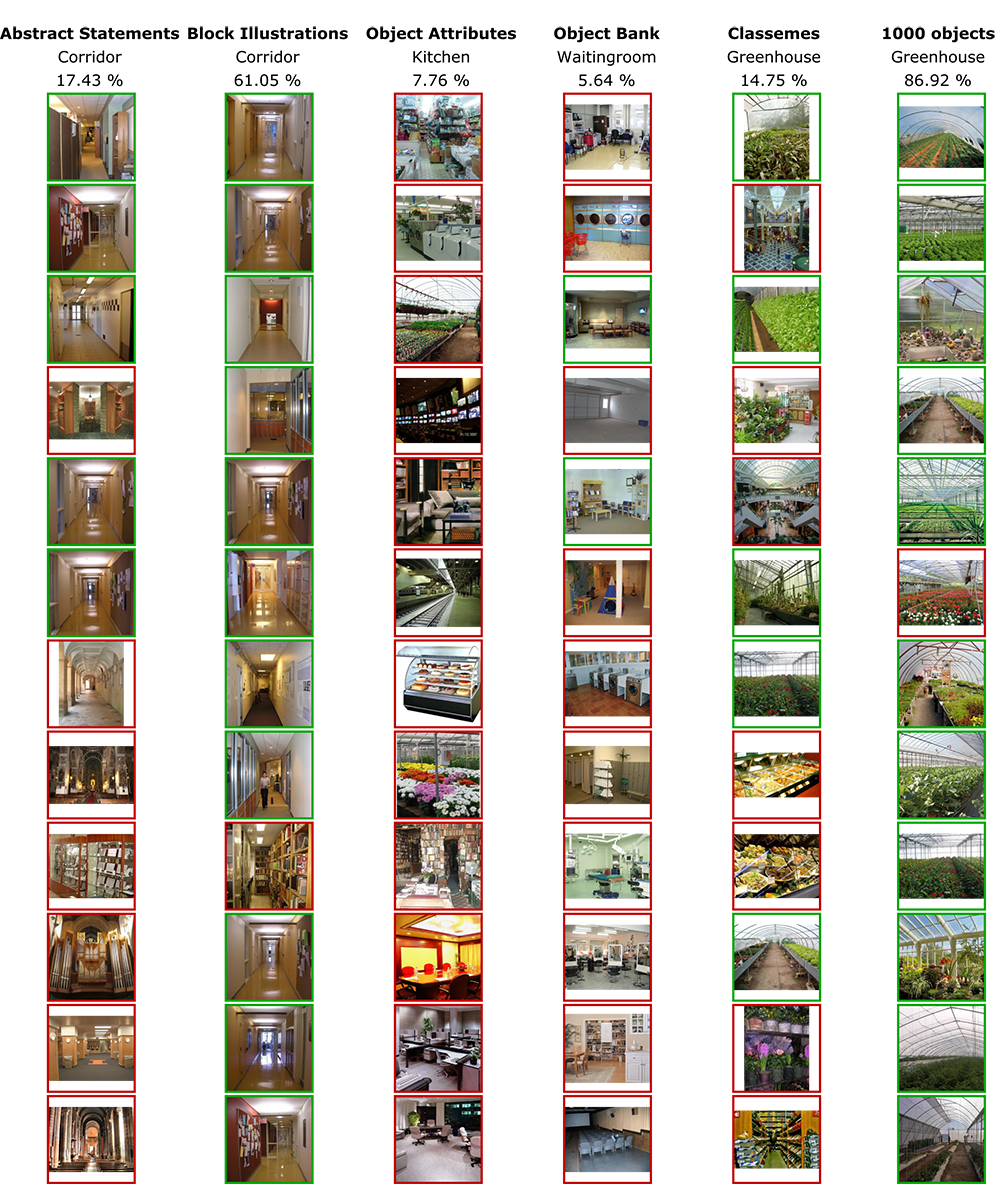}
\caption{ Top retrieved images for the best performing scene of
six methods on Indoor67, and their average precision.
Interestingly, since the abstract statements and block
illustrations capture the properties of things layout, the best
performing scene is \emph{corridor}, which indeed has a specific
things layout. The semantic methods use appearance features and
their best performing scene depends on the performance of the
semantic classes within. We conclude, the things syntax captures
the scene things layout, and it can be used to retrieve scenes
with abstract descriptions which require no learning at all. }
\label{fig:RetrievalExamples}
\end{figure*}

In the last experiment we compare our abstract representations,
query by statements and query by block illustrations, to four
other representations involving semantic descriptions of object
identities \cite{objbankIJCV13}, object attributes
\cite{farhadi2009describing}, general classemes categories trained
from Flickr \cite{classemesPAMI} and 1000 ImageNet objects
\cite{ILSVRC15}. Each method has different number of semantic
classes, object bank with 177, object attributes with 67,
classemes with 2,659, and ImageNet with 1,000 semantic classes.
For all methods we use the available software online, provided by
the authors, to get semantic scores representation on the test
images. For ImageNet we get scores from the output layer of 1,000
classes from an in-house developed Convolutional Neural Network
inspired by \cite{DeepNetEccvZeilerF14} and trained on ImageNet.
We use the same principle for all semantic methods to create a
semantic representation per scene to query by. We rely on object
annotations from LabelMe~\cite{LabelMeRussell2008} on both
datasets, Indoor67 and SUNAttributes. For example, classemes has
2,659 semantic classes. We count how often a semantic class from
classemes was annotated by humans in images from a scene class.
This gives us a distribution histogram of the classemes objects
per scene, which we L1 normalize and use as a scene representation
to query by. On a test image we run the classemes software as
provided by the authors, and get classemes scores. To calculate a
ranking score of a test image for a scene, we use the DAP model as
proposed by~\cite{lampert2013attribute} between the classemes
scene representation and the classemes scores of the test image.
The same procedure we repeat for all semantic methods. We realize
we use these methods in a different way than they were originally
intended to be used. The reason we choose these works is because
the semantic information they provide is not directly aimed to
describe scenes, like the case of provided scene attributes in
\cite{lampert2013attribute}. Our goal is to show that when
semantic information is used, it is beneficial to correlate the
semantics to the category of the test classes, like annotations of
animal attributes for animal classes \cite{Lampert09learningto},
or scene attributes for scene classes \cite{lampert2013attribute}.
We believe that when no direct semantic information to the
category of the test classes is available, abstract descriptions
are a good alternative, since they generalize well to any unseen
category. We show this on scene categories.

We present mean average precision results in
Table~\ref{table:ZeroShotResults}. Since the object attributes
\cite{farhadi2009describing} are intended for a different purpose,
they do not generalize well to scenes. Object bank
\cite{objbankIJCV13} is successful when one of its objects has an
(accidental) relationship with the scene of interest.
Unfortunately, most scenes have a low average precision as an
overlap with the objects in the bank is missing. Classemes
\cite{classemesPAMI} detectors are trained from weakly supervised
images, but their representation is more rich, having over 2K
semantic classes, and results in better MAP in general. Similar to
object bank, if the objects in the 1,000 ImageNet objects overlap
with the scenes, it results in good average precision, else it
fails. In a scenario where there is no semantic information for
the intended category available, our abstract representation is
more general, leading to reasonable accuracy values in almost all
settings. This can be clearly seen in
Figure~\ref{fig:ConfussionMatricesAbstVsSens}, where we show
confusion matrices on the Indoor67 dataset of our abstract
representations and other semantic representations. When there is
a good overlap of the scene class with the object, like for
example \emph{closet} which is an object class in ImageNet and a
scene class in Indoor67, then there is a strong response as seen
in the red lines in Figure~\ref{fig:ConfussionMatricesAbstVsSens}.
We show top retrieved images of the best performing scene for all
methods on the Indoor67 dataset in
Figure~\ref{fig:RetrievalExamples}. Interestingly, since the
abstract statements and block illustrations capture the properties
of things layout, the best performing scene is \emph{corridor},
which indeed has a specific things layout. The other methods use
appearance features, and the best performing scene depends on the
performance of their semantic classes. Overall we conclude that
when there is no direct semantic information available for unseen
classes in a scene retrieval setting, abstract descriptions of
things syntax are a good alternative. An added benefit of our work
is that it does not require any labeled examples to build the
representation. The abstract things syntax is directly generated
from the image itself.

\section{Conclusions}

In this paper we open up further understanding of the rules
composing the visual world around us, the potential to combine the
objects layout information, and the recognition of scenes without
the need to keep semantics like object types or attributes. We
show that next to object types, there is another source of
information defining what makes a scene. Object types are unknown
in abstract paintings, architecture inspiration, microscopic and
cosmic observation, while their ``things'' composition is well
observable. Thus, to describe images of these compositions in an
abstract manner is inevitable. We give preference to the
composition of ``things'' as an indicator for the type of scene.
We start from ``things'' as defined by modern object proposals,
and investigate their immediately observable features: position,
size, aspect ratio and color. We name the ensemble of things
properties as things syntax, and we investigate its effectiveness
to represent and retrieve a scene. From four experiments we
conclude, (1) the distribution of thing-features from proposal
methods approximates the distribution of thing-features from human
annotated things closely. We investigate and analyze the
discriminative potential and properties of the things syntax when
translated into (2) abstract language statements and (3) abstract
block illustrations for scene retrieval. In both cases we show
that things aspect ratio is the most informative property, also
that scenes can still be retrieved if their things deviate up to 8
pixels from the true objects, and by providing more precise
abstract statements, \ie more bins in the histogram
representations, or by using more GMM prototypes in the Fisher
vector representations of block illustrations, the retrieval
results on fine-grained scenes improves. At last, (4) we compare
the abstract things syntax representations with four other
semantic representations which are not directly aimed for scenes.
We show that when there is an accidental overlap of the semantic
classes with the scene class, using semantics is beneficial.
However, when there is no accidental overlap, the abstract
descriptions of things syntax are a good alternative, and in some
cases show even better retrieval performance than the semantic
representations which require learning beforehand. Overall and
surprisingly, we show that even though we use the simplest of
features from things layout, we can still retrieve scenes
reasonably well, and with an additional benefit that we do not
require any learning examples.

\vspace{0.5cm}

\textbf{Acknowledgements}

\vspace{0.2cm}

This research is supported by the STW STORY project and the Dutch
national program COMMIT.










\bibliographystyle{elsarticle-num-names}
\bibliography{boxStatistics}

\end{document}